\documentclass[preprint,12pt]{elsarticle}

\usepackage{amssymb}
\usepackage{amsmath}
\usepackage{multirow}
\usepackage{graphicx}
\usepackage{amssymb}
\usepackage{pifont} 
\usepackage[table]{xcolor}
\usepackage{hyperref}

\journal{Nuclear Physics B}

\begin{document}

\begin{frontmatter}



\title{A Joint Learning Framework with Feature Reconstruction and Prediction for Incomplete Satellite Image Time Series in Agricultural Semantic Segmentation}


\author[1]{Yuze Wang}
\author[2]{Mariana Belgiu}
\author[1]{Haiyang Wu}
\author[1]{Dandan Zhong}
\author[2]{Yangyang Cao}
\author[1]{Chao Tao$^*$}

\affiliation[1]{organization={School of Geosciences and Info-Physics, Central South University},
            addressline={No. 932, Lushan Nan Road}, 
            city={Changsha},
            postcode={410083}, 
            country={China}}
\affiliation[2]{organization={Faculty of Geo-Information Science and Earth Observation (ITC), University of Twente},
            addressline={PO BOX 217}, 
            city={Enschede},
            postcode={7500 AE}, 
            country={The Netherlands}}

\begin{abstract}
Satellite Image Time Series (SITS) is crucial for semantic segmentation in agriculture. However, Cloud contamination introduces time gaps in SITS, disrupting temporal dependencies and causing feature shifts, leading to degraded performance of models trained on complete SITS. Existing methods typically address this by reconstructing the entire SITS before prediction or using data augmentation to simulate missing data. Yet, full reconstruction may introduce noise and redundancy, while the data-augmented model can only handle limited missing patterns, leading to poor generalization. We propose a joint learning framework with feature reconstruction and prediction to address incomplete SITS more effectively. During training, we simulate data-missing scenarios using temporal masks. The two tasks are guided by both ground-truth labels and the teacher model trained on complete SITS. The prediction task constrains the model from selectively reconstructing critical features from masked inputs that align with the teacher’s temporal feature representations. It reduces unnecessary reconstruction and limits noise propagation. By integrating reconstructed features into the prediction task, the model avoids learning shortcuts and maintains its ability to handle varied missing patterns and complete SITS. Experiments on SITS from Hunan Province, Western France, and Catalonia show that our method improves mean F1-scores by 6.93\% in cropland extraction and 7.09\% in crop classification over baselines. It also generalizes well across satellite sensors, including Sentinel-2 and PlanetScope, under varying temporal missing rates and model backbones. The source codes are available at: \url{https://github.com/wangyuze-csu/Joint_FRP}
\end{abstract}



\end{frontmatter}



\section{Introduction}
\label{sec1}

In recent years, Deep Learning (DL) methods leveraging remote sensing (RS) images have played an important role in automatic agricultural monitoring tasks. Among them, semantic segmentation has shown significant potential in tasks such as cropland extraction and crop classification \cite{Zhang_GeneralizedApproachBased_2020,Wang_EvaluationDeeplearningModel_2022,wu2025fsvlm}. These applications play a critical role in supporting agricultural resource management \cite{Jiang_MethodMappingRice_2018}, policy planning \cite{perselloDelineationAgriculturalFields2019}, and sustainable development \cite{Gebbers_PrecisionAgricultureFood_2010}. With the increasing availability of high temporal resolution RS images, dense Satellite Image Time Series (SITS) have become a valuable source of information for agricultural semantic segmentation tasks\cite{Xu_InterpretingMultitemporalDeep_2021,Waldner_AutomatedAnnualCropland_2015}. The detailed temporal dynamics captured by SITS allow DL models to effectively learn and exploit the phenological patterns of different vegetation \cite{Waldner_AutomatedAnnualCropland_2015}. This temporal information enhances the class separability not only between different crop types but also between cropland and other land covers \cite{Belgiu_Sentinel2CroplandMapping_2018}.

However, in practical application scenarios, ensuring the completeness and quality of collected SITS remains challenging due to frequent cloud contamination. According to the International Satellite Cloud Climatology Project-Flux Data (ISCCP-FD), the global monthly cloud cover is approximately 66.5\% \cite{Zhang_CalculationRadiativeFluxes_2004,mao2019changes}. Such pervasive cloud presence often results in missing temporal observations, which not only disrupts the temporal dependencies across time steps but also induces a shift in the overall distribution of temporal information. Consequently, models trained on complete SITS often struggle to generalize to incomplete sequences, as they fail to capture the underlying phenological dynamics present in real-world data \cite{Gu_ExplicitHazeCloud_2022,Wu_ProgressiveGapfillingOptical_2024}. This impairs the model's ability to leverage temporal information for accurate and robust prediction.

For images with minor cloud contamination, such as thin clouds or small areas of thick cloud cover, cloud removal methods \cite{Li_CloudCloudShadow_2022,Ajayi_PerformanceEvaluationSelected_2022} can successfully reconstruct the affected areas, thereby preserving the temporal completeness of SITS. However, persistent and severe cloud contamination presents significant challenges to these methods. Methods that depend on multi-temporal imaging\cite{Wang_UnsupervisedDomainFactorization_2023}, typically require the availability of cloud-free reference images within a relevant temporal window, which is often impractical under such conditions. Alternative approaches attempt to reconstruct cloud-covered regions using spatial and spectral information from cloud-free areas within a single image \cite{Tao_ThickCloudRemoval_2022}. Yet, when cloud coverage exceeds a critical threshold (e.g., $>$ 50\% coverage), the significant loss of contextual information makes it hard for these methods to recover the missing areas accurately. These heavily or fully cloud-obscured images introduce time gaps in SITS. As a result, maintaining the performance of deep learning models with incomplete SITS has become a critical and unresolved challenge in agricultural applications. 

Recently, two primary strategies have been developed to address the challenges posed by incomplete SITS: Data Reconstruction (DR) and Data Augmentation (DA). DR methods aim to reconstruct the SITS before the input into multi-temporal semantic segmentation models (hereafter referred to as the prediction models) \cite{Shen_MissingInformationReconstruction_2015}. One category of DR methods makes use of the external information from other optical or Synthetic Aperture Radar (SAR) sensors to fill the time gaps \cite{Pipia_FusingOpticalSAR_2019}. However, acquiring cloud-free optical images within the relevant temporal window is often impractical due to the persistent nature of cloud cover. The acquisition and processing of SAR data increase time and manual costs \cite{Tsokas_SARDataApplications_2022a}. Furthermore, integrating SAR and incomplete optical SITS under low-cost remains a challenge due to their distinct sensing modalities \cite{Yuan_BridgingOpticalSAR_2023}. Another type of DR method is using the internal information from the incomplete STIS itself. These methods typically rely on interpolation \cite{Lepot_InterpolationTimeSeries_2017} or Deep learning(DL)-based methods \cite{Moskolai_ApplicationDeepLearning_2021}. Although the interpolation might perform well when the time gaps are sparsely distributed in SITS, it fails in cases of extensive or continuous time gaps\cite{Lepot_InterpolationTimeSeries_2017, Li_AnomalyDetectionTime_2021}. DL-based reconstruction can offer improved results, but may still contain noise and distort temporal consistency \cite{Quan_DeepLearningBasedImage_2024, Fang_TimeSeriesData_2020}. More critically, complete reconstruction is redundant for prediction tasks. Prediction models primarily depend on key temporal patterns rather than the entire time series \cite{Gangopadhyay_SpatiotemporalAttentionMultivariate_2021, Cai_RevisitingEncodingSatellite_2023}. Thus, complete reconstruction might not only involve unnecessary computational and storage costs, but also introduce the risk of overfitting to noise and irrelevant information \cite{Casolaro_DeepLearningTime_2023}.

The DA methods aim to enhance the model's robustness to incomplete SITS by increasing the diversity of training data through simulated temporal missing conditions \cite{wen2020time}. In the temporal domain, two widely adopted strategies in RS are Window Slicing (WS) and temporal Dropping (TD) \cite{Gao_DataAugmentationTimeSeries_2024}. WS augments the training data by sampling continuous local temporal segments, enabling models to infer results using localized phenological cues \cite{Guennec_DataAugmentationTime_2016}. TD, on the other hand, randomly removes time steps across the full sequence, enhancing the model's ability to handle globally discontinuous temporal patterns. Although these strategies are simple and effective, they face limitations. These include the risk of catastrophic forgetting \cite{Kirkpatrick_OvercomingCatastrophicForgetting_2017}, where the model loses its ability to process complete SITS, and mutual interference when learning from diverse missing patterns \cite{schak2019study}. Moreover, models trained under simulated temporal missing conditions may suffer from the shortest-path phenomenon \cite{geirhos2020shortcut}, where they learn shortcut reasoning strategies rather than generalizable temporal representations. The model tends to rely on the shortest and computationally least expensive temporal jump connections, while neglecting more complete but longer temporal dependency chains \cite{geirhos2020shortcut}. It will compromise the model’s generalizability to unseen data conditions.

To address these problems, we proposed a joint learning framework with feature reconstruction and prediction for multi-temporal agricultural segmentation tasks. It enables the model to selectively reconstruct essential temporal features that are beneficial for prediction under temporal missing conditions. Specifically, we simulate incomplete SITS by randomly masking time steps during training. The model is jointly guided by both prediction labels and a teacher model trained on complete SITS. The feature reconstruction task encourages the model to generate temporal features from masked inputs that approximate those extracted by the teacher model from complete SITS. Simultaneously, supervision from prediction labels ensures that the model learns to recover task-relevant temporal features rather than reconstructing irrelevant or noisy information. By focusing on selective reconstruction, the method reduces both the computational/storage costs for complete reconstruction of the SITS. It also mitigates the risk of error propagation from noise that may be contained in the reconstructed SITS. For the prediction tasks, the reconstructed temporal feature can support the reasoning process. This helps the model avoid shortcut reasoning and enhancing its ability to capture long-term temporal dependencies. Additionally, the teacher model is also used to constrain the model’s prediction process, ensuring that the model maintains its capability to handle the complete SITS. 

In summary, feature reconstruction prevents the model from acquiring incorrect reasoning ability during prediction, whereas prediction maintains reliability and reduces the redundancy of reconstructed information. To validate our framework, we conduct experiments on cropland extraction and crop classification tasks with SITS. It contains three study areas in Hunan Province and Western France \& Catalonia , using images from Sentinel-2 and Planet satellites, respectively. We also assessed the generalization ability of the proposed framework in varied model architectures and investigated its internal mechanisms.
In summary, our main contributions are as follows:

\begin{enumerate}
	\setlength\itemsep{0em}\setlength\parskip{0em}\setlength\topsep{0em}\setlength\partopsep{0em}\setlength\parsep{0em} 
	\item{We proposed a joint leanring framework with feature reconstruction and prediction for the multi-temporal agricultural semantic segmentation task in severe temporal missing conditions. It outperforms the state-of-the-art method in cropland extraction and crop classification tasks.} 
        \vspace{5pt}
	\item{The proposed framework demonstrates strong generalization across different sensors, regions, and varied numbers of time gaps, while being agnostic in various model backbones. It significantly improves model robustness for real-world agricultural monitoring.}
        \vspace{5pt}

\end{enumerate}

The remainder of this paper is organized as follows: Section \ref{sec2} reviews related work on data reconstruction and data augmentation methods. Section \ref{sec3} describes the data processing pipeline and the proposed framework in detail. Section \ref{sec4} presents the experimental setup, comparative results with other methods, and the model's performance under different numbers of time gaps. Section \ref{sec5} discusses the model-agnostic nature and internal mechanisms of the proposed approach. Finally, Section \ref{sec6} concludes the paper and outlines directions for future work.

\section{Related work}
\label{sec2}

\subsection{The data reconstruction for time gaps in SITS}
\label{subsec2.1}

Data reconstruction aims to impute the time gaps in SITS, thereby providing complete temporal information to support the semantic segmentation models in prediction tasks \cite{Shen_MissingInformationReconstruction_2015}. Depending on the source of supplementary information, these methods can be broadly categorized into external and internal imputations. The external imputation typically fills the data gaps by utilizing the information from other optical or SAR sensors \cite{Luo_STAIRGenericFullyautomated_2018, Cresson_OpticalImageGap_2019}. According to the varied revisit cycles of different satellites, the optical-based method uses adjacent available images from other satellites to complete the SITS \cite{Luo_STAIRGenericFullyautomated_2018}. However, acquiring such images within a relevant temporal window is often impractical, particularly in regions with persistent cloud cover during rainy seasons, such as tropical, monsoon, and maritime climates. SAR images have the ability to penetrate clouds, making them a reliable source of information in cloudy scenarios. Nevertheless, SAR data acquisition and preprocessing can be time-consuming \cite{Passah_SARImageClassification_2022}. Moreover, due to the distinct imaging mechanisms of optical and SAR images, different statistical properties, temporal and semantic patterns make it challenging to directly use SAR to fill time gaps in optical SITS. Although \cite{Cresson_OpticalImageGap_2019}and \cite{Yuan_BridgingOpticalSAR_2023} learn the mapping relationships between SAR and optical data to transform SAR information, thereby jointly constructing complete time information. Considering the diverse characteristics of optical and SAR images that are exhibited in different locations and across years \cite{Huang_ClassificationLargeScaleHighResolution_2021a}, they still need extensive paired optical-SAR samples to relearn the cross-modality relationships for different scenarios. It still increased the computational costs and data requirements.

To address the cost issue, the internal imputation methods typically leverage available temporal information within the SITS to fill the time gaps \cite{Shen_MissingInformationReconstruction_2015}. Among them, interpolation estimates missing values under the assumption of temporal continuity, using neighboring time points as references \cite{Lepot_InterpolationTimeSeries_2017}. For instance, linear interpolation (Linear-IN), closest interpolation (Closest-IN), and last interpolation (Last-IN) are widely used in the RS field. However, these methods are challenged when faced with long or continuous temporal gaps, as the limited contextual information often leads to substantial deviations from true values \cite{Lepot_InterpolationTimeSeries_2017}. With the advancement of deep learning methods, data-driven models can learn the changing pattern of temporal features to fill the time gaps in incomplete SITS \cite{Moskolai_ApplicationDeepLearning_2021}. These methods are usually trained on simulated incomplete SITS, with the optimization goal of generating data that closely approximates complete SITS. For example, GANs (Generative Adversarial Networks) \cite{gonzalez2025generative} employ an adversarial strategy between the generator and the discriminator, which forces the generators to estimate the time gaps that are indistinguishable from complete SITS by the discriminator. TSVit (Time Series Vision Transformer) \cite{Yuan_SITSFormerPretrainedSpatiospectraltemporal_2022} captures global temporal dependencies through self-attention mechanisms and utilizes cross-attention to select information relevant to the missing information, thereby reconstructing the time gaps. Additionally, the integration of convolutional and pooling operations with attention mechanisms enables more comprehensive extraction of multi-scale local temporal features, enhancing the modeling of global temporal dynamics \cite{Stucker_UTILISESequencetoSequenceModel_2023}. Despite these advances, reconstructing large or continuous gaps remains challenging. The severe lack of contextual information may cause the reconstructed data to deviate from the overall distribution of SITS, thereby introducing noise \cite{Quan_DeepLearningBasedImage_2024, Fang_TimeSeriesData_2020}. The propagation of such noises may further affect the reasoning process of prediction models, resulting in unreliable segmentation results. 

In summary, the external imputation methods are limited by the availability of optical data and the cost of using SAR data. The internal imputation methods are unsuitable for handling the extensive or continuous time gaps in SITS. Even with the support of deep learning reconstruction methods, it still faces the risk of noise propagation. More importantly, fully reconstructing all missing information is often unnecessary for multi-temporal semantic segmentation tasks. Although the reasoning process of the model’s prediction jointly relies on spatial and temporal features, spatial information in SITS is often redundant [27]. In the temporal dimension, the model mainly relies on the key temporal patterns extracted from SITS [26]. Reconstructing the entire sequence may introduce irrelevant information, increasing computational/storage demands and risking overfitting to noise or non-essential features.

\subsection{The data augmentation for time gaps in SITS}
\label{subsec2.2}

Data augmentation is widely used in the RS field to enhance the robustness of models in complex scenarios by manually increasing the diversity of training data. For the multi-temporal segmentation tasks, exposing the model to scenarios with missing temporal data during training has proven effective for improving its ability to handle time gaps during prediction \cite{Gao_DataAugmentationTimeSeries_2024}. Among them, the most commonly used methods are window slicing (WS) and temporal dropout (TD) \cite{Gao_DataAugmentationTimeSeries_2024, Yuan_EmpiricalStudyData_2025a}. The WS strategy refers to selecting a temporal window (i.e., a continuous sequence of time steps) from the original SITS to simulate scenarios with missing data \cite{Kirkpatrick_OvercomingCatastrophicForgetting_2017}. WS encourages the model to focus on local temporal features and to infer predictions based on partial sequences. However, the WS may cause the model to pay more attention to local features, weakening its capacity to capture global temporal dependencies. TP is a more general strategy that typically removes time steps either individually or continuously during training to simulate temporal missing scenarios \cite{deAlbuquerque_DealingCloudsSeasonal_2021}. It enables the model to infer complete temporal features from partially observed data, thereby reducing the model’s reliance on patterns in complete SITS. Furthermore, the TP can prevent the model from over-relying on specific time steps and guide it to learn more robust and generalized temporal features. 
For models handling incomplete SITS, the customized data augmentation strategy is considered beneficial for enhancing the model's robustness under specific time-missing scenarios \cite{iglesias2023data}. Since the distribution of missing data in SITS is often influenced by local climate conditions, selective TD can be effective in region-specific scenarios. However, customizing specific strategies for every region with different climate conditions and employing a customized model is a time-consuming process \cite{iglesias2023data}. Additionally, while cloud contamination often follows predictable seasonal trends, it also contains inherent randomness. The fixed TP strategy may cause the model to lose its ability to handle exceptional cases. For example, in northeastern Hunan, although Sentinel-2 data show a high cloud-related missing rate in summer and autumn (averaging 50\%), non-negligible gaps also occur in spring and winter (averaging 15\%). Therefore, the random TP strategy became an efficient and robust strategy in the RS field.

Although the data augmentation is an effective way to enhance the model's ability to handle the time gaps in the multi-temporal segmentation task, it still faces several challenges. First, due to catastrophic forgetting \cite{Kirkpatrick_OvercomingCatastrophicForgetting_2017}, models with limited memory capacity may overfit incomplete SITS, thereby diminishing their ability to process complete SITS effectively \cite{schak2019study}. Moreover, training on multiple temporal missing patterns may result in mutual interference, making it difficult for the model to generalize across varied missing conditions. Second, during the prediction of segmentation tasks, the model may fail to develop appropriate reasoning capabilities due to the shortcut learning phenomenon \cite{geirhos2020shortcut}. The model tends to take the easier and computationally least expensive way to predict the final result. The model may overlook the intermediate dynamic modeling process within the full temporal dependency chain and instead adopt jump connections for reasoning. For instance, the model may learn to infer crop types directly by relying only on observations from January and April in simulated temporal missing scenarios, while ignoring the intermediate growth stages. As a result, in a real-world scenario, even when data from February and March becomes available, the model continues to follow this shortcut path instead of incorporating it into a complete temporal reasoning chain. Once the model develops a reliance on such shortcut behavior, it tends to learn the correlation between missing patterns and the final prediction result, rather than building a global representation from limited observations. This not only impairs the model's ability to capture long-term temporal dependencies, but also degrades its generalization to unseen data.

\section{Materials and methods}
\label{sec3}

\subsection{Preprocessing and datasets}
\label{subsec3.1}

Our study areas for dataset construction include Hunan Province in China and two European regions: Western France and Catalonia. Hunan Province has a continental subtropical monsoonal humid climate, while the European regions have temperate maritime and Mediterranean climates. All study areas have frequent and persistent cloud cover, which might lead to extensive and continuous time gaps in the acquired SITS. For China's Hunan Province, we prepared cropland extraction datasets using two sensors with different spatial and temporal resolutions: Sentinel-2 and PlanetScope constellation, resulting in the Hunan SEN and Hunan PLA datasets, respectively. For the European regions, we constructed the Fr\&Cat S4A dataset based on the S4A for crop classification \cite{Sykas_Sentinel2MultiyearMulticountry_2022a}. We further divided each dataset into complete and incomplete subsets based on the temporal availability of SITS. The complete subsets (60\% for training and 40\% for testing) were used for simulated experiments, and the incomplete subsets were only used for real-world testing. Detailed information and processing steps are presented below.

\subsubsection{Hunan SEN}
\label{subsec3.1.1}
We collected Sentinel-2A and Sentinel-2B Level-2A Bottom of Atmosphere reflectance images (S2 L2A) from Google Cloud Platform, covering the period from January 2019 to December 2019. Bands with 10m and 20m spatial resolutions were used for dataset construction, with the 20m bands resampled to 10m resolution. Furthermore, we used the Quality Assurance (QA) bands provided by the ESA on Google Earth Engine (GEE) as a reference to calculate the cloud covers \cite{amani2020google}. 

\begin{figure}[t]
	\begin{center}
        \includegraphics[width=1\linewidth]{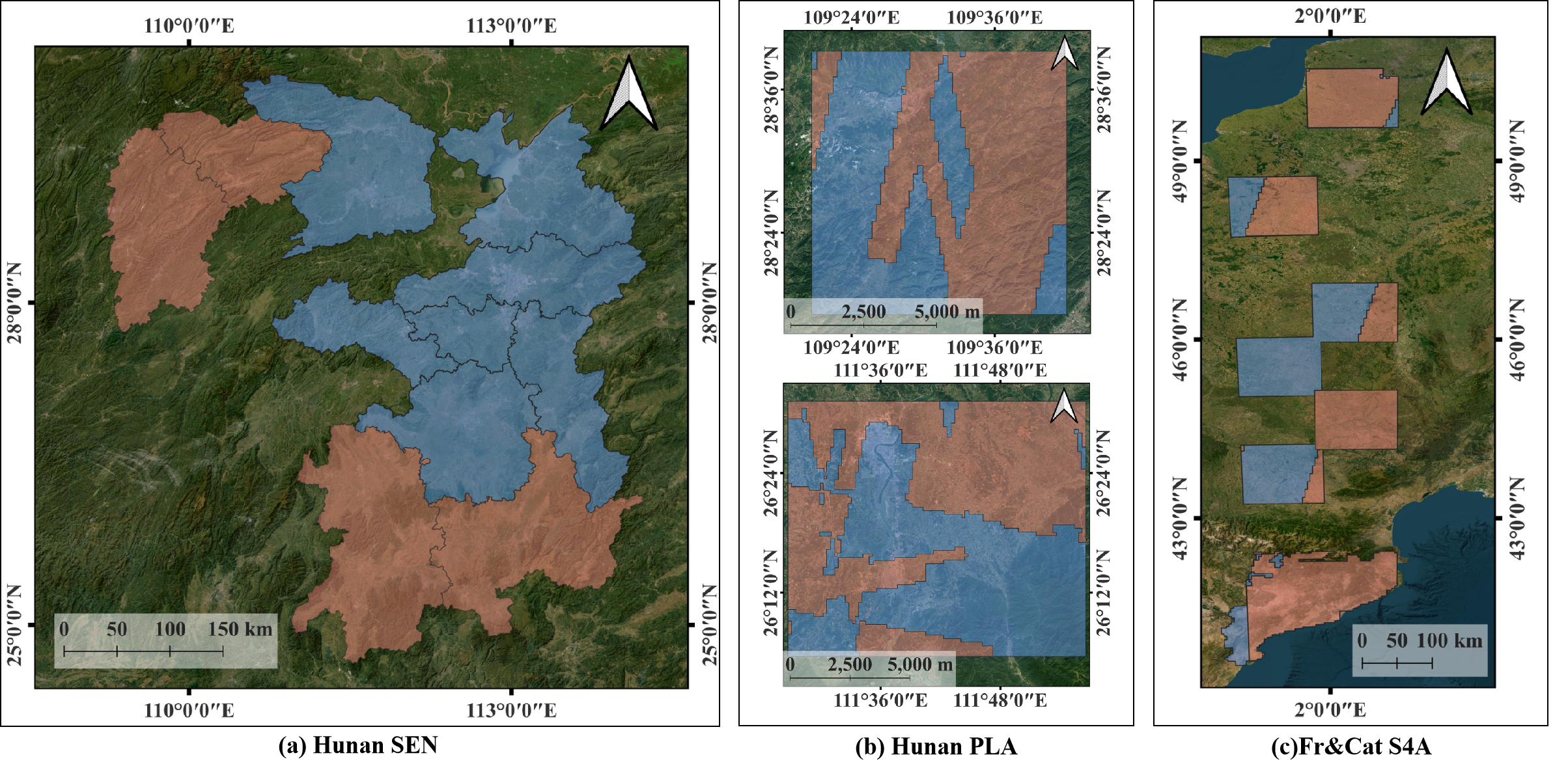}
		\caption{The location and extent of datasets, where red and blue regions represent complete and incomplete subsets, respectively.}
		\label{F1}
	\end{center}
\end{figure}

\begin{figure}[t]
	\begin{center}
        \includegraphics[width=1\linewidth]{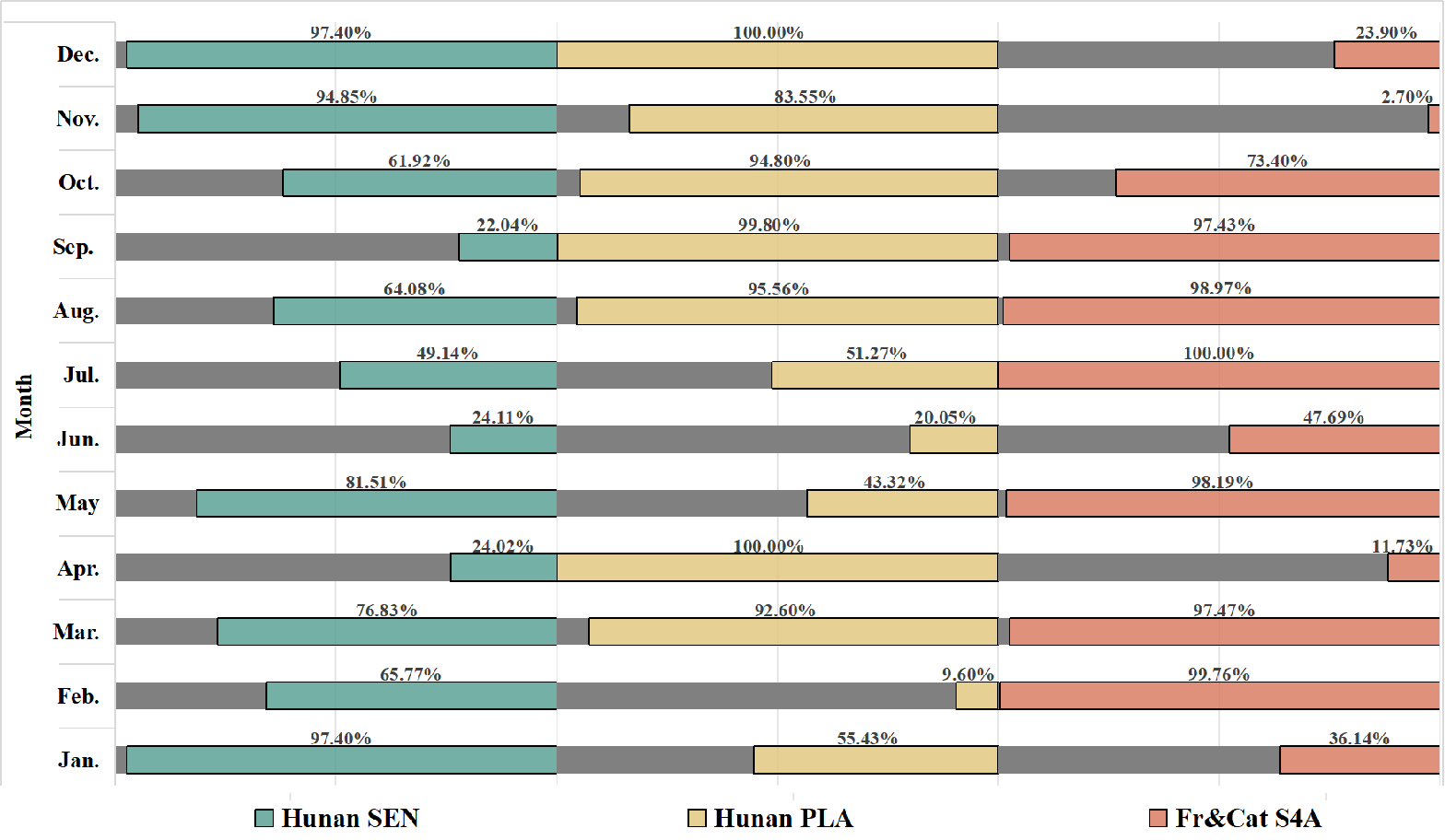}
		\caption{Monthly data availability of SITS samples in incomplete subsets: Hunan SEN, Hunan PLA, and Fr\&Cat S4A datasets}
		\label{F2}
	\end{center}
\end{figure}

As shown in Fig. \ref{F1}(a), for the complete subset, we randomly sampled 256×256-pixel sub-regions across seven city-level administrative areas of northeastern Hunan, where cloud cover frequency was relatively low. For each sub-region, cloud-covered pixels were excluded to generate monthly median composite images from all available pixels. Composite images with cloud-cover ratios $<$ 20\% are treated as clean images and used to assemble SITS. After filtering, 5000 SITS samples with complete 12-month images were obtained. For the incomplete subset, we randomly sampled 256×256-pixel sub-regions across four city-level administrative areas of northwestern and southeastern Hunan, where cloud cover frequency was relatively high. Monthly composite images with cloud cover $>$ 40\% were treated as missing. The remaining cloud-cover pixels were filled using linear interpolation from the nearest available observations. After filtering, 3,048 SITS samples with 4-9 available monthly images were obtained. The monthly image availability for incomplete samples is shown in Fig. \ref{F2}. For all samples, the labels are obtained by visually interpreting the RGB bands of Sentinel-2 images, assisted by high-resolution Google images.

\subsubsection{Hunan PLA}
\label{subsec3.1.2}

We acquired PlanetScope-2 (PS2) images from the Planet Platform, covering the period from January 2019 to December 2019. They include Blue, Green, Red, and Near-Infrared spectral bands with a spatial resolution of 3.7 m. As shown in Fig. \ref{F1}(b), the regions are located in the Jishou and Yongzhou areas. During the downloading, we selected three images with the lowest cloud-cover ratios for the target area, prioritizing an even distribution throughout each month. We also manually masked the cloud-cover regions for each image and generated monthly median composite images.

All images were cropped to 256×256-pixel images, and sub-images with $>$ 40\% cloud-covered ratios were treated as missing. The remaining cloud-cover pixels were filled using linear interpolation from the nearest available observations. After assembling monthly images into SITS samples, we divide them into complete and incomplete subsets according to the available time length. The complete subset contained 5184 samples with 10-12 monthly images, where samples missing 1-2 monthly images were gap-filled using linear interpolation from the nearest available images. The incomplete subset contained 4584 samples with 4-9 available monthly images. The monthly image availability for incomplete samples is shown in Fig. \ref{F2}. For all samples, the labels are obtained by visually interpreting the RGB bands of PS2 images.

\subsubsection{Fr\&Cat S4A}
\label{subsec3.1.3}

Based on S4A datasets, we extracted data from January to December 2019, including 10 spectral bands at both 10m and 20m spatial resolutions. The 20m resolution bands were resampled to 10m to match the other bands. Each SITS sample covered a 366×366-pixel area, with the cloud cover ratio of every image calculated using the provided Hollstein masks. Images with $>$ 40\% cloud-cover pixels were treated as missing, followed by additional manual filtering of remaining unusable images. We created month median composites with the remaining invalid pixels filled via linear interpolation from the nearest valid observations.

The complete subset obtained 4771 samples with 10-12 monthly images, where samples missing 1-2 monthly images were gap-filled using linear interpolation from the nearest available images. The incomplete subset contained 2925 samples with 4-9 monthly images. The monthly image availability for incomplete samples is shown in Fig. \ref{F2}. For all samples, we combined the S4A labels into 8 categories: Background (BG), Cereals (Cer), Vegetables \& Melons (V\&M), Root and Tuber crops (R\&T), Leguminous Crops (LC), Oilseed Crops (OC), Sugar Crops (SC), and Other Crops (OT).

\subsection{Method}
\label{subsec3.2}

Multi-temporal semantic segmentation models are designed to process time-series inputs and produce a single semantic output, widely applied in agricultural tasks such as cropland extraction and crop classification. The input Satellite Image Time Series (SITS) is typically represented as $X\ \in\mathbb{R}^{T\times C\times H\times W}$, and the model extracts spatial-temporal features to generate a pixel-wise one-hot prediction $Y\ \in\mathbb{R}^{H\times W\times K}$ with K classes. 

As shown in Fig. \ref{F3}, our proposed framework aims to jointly optimize the model through feature reconstruction and prediction tasks under simulated data-missing scenarios, thereby enhancing its ability to handle incomplete SITS. These two tasks are simultaneously constrained by the teacher model pre-trained in complete SITS and the ground truth labels. (1) In the feature reconstruction task, the teacher model serves as a reference to demonstrate to the student model what the ideal spatial-temporal features extracted from complete SITS should look like. Accordingly, the student model is encouraged to reconstruct features from incomplete SITS that align with the ideal features. Meanwhile, ground truth labels also encourage the model to selectively reconstruct the key patterns of features that are beneficial for predictions, which balances the trade-off between reconstructed feature completeness and prediction reliability. (2) In the prediction task, supported by these reconstructed features, the student model was encouraged to imitate the teacher’s decision-making process. The model is supervised by both the teacher's logit outputs (from complete SITS) and ground truth labels, enabling it to learn the intrinsic inter-class relationships and underlying reasoning logic for robust feature representation. This strategy prevents the student model from overfitting to limited temporal missing patterns and maintains its ability to generalize to complete sequences. The detailed workflow is illustrated as follows:

\begin{figure}[t]
	\begin{center}
        \includegraphics[width=1\linewidth]{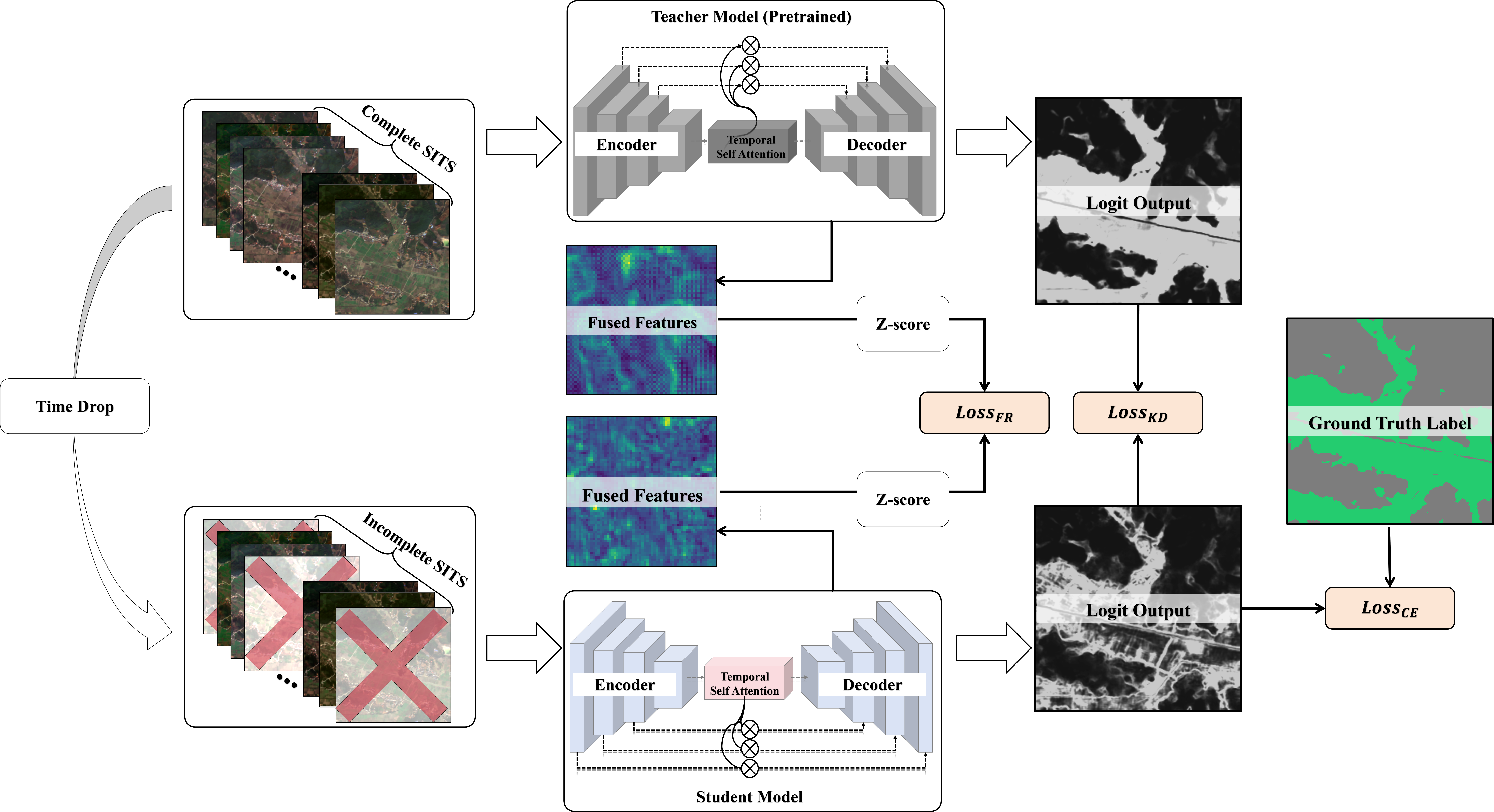}
		\caption{The general framework of the proposed method}
		\label{F3}
	\end{center}
\end{figure}

Firstly, we conduct teacher model pre-training and simulate data-missing scenarios. We define a training dataset $\mathcal{D} = \{X_i\}_{i=1}^N$, where each sample $X_i \in \mathbb{R}^{T \times C \times H \times W}$ represents a complete Satellite Image Time Series (SITS). The teacher model $f_T$ is pre-trained on $\mathcal{D}$ and is frozen afterward for guiding the student model. For every sample $X \in \mathcal{D}$, we generate an incomplete version $X^\prime \in \mathbb{R}^{T^\prime \times C \times H \times W}$ (where $T^\prime < T$) through random temporal masking:

\begin{equation}\label{equ1}
    \mathbf{X}' = \mathrm{TemporalMask}(\mathbf{X}, r), \quad r \sim \mathcal{U}(M, N)
\end{equation}
where $M$ and $N$ denote the minimum and maximum mask ratios in the temporal dimension, respectively. Considering real-world temporal missing rates, $M$ and $N$ are typically set to 25

Secondly, we guide the student model $f_S$ to selectively reconstruct critical features from Incomplete SITS. The ${X}$ and $X^\prime$ are respectively fed into the $f_T$ and $f_S$ to produce spatial-temporal features $F$ and $F^\prime$. Considering the redundancy of spatial information in SITS, we use the deepest temporally fused feature as a supervision target. This encourages the student model to focus on reconstructing global temporal dependencies and dynamics, rather than spatial patterns. For instance, in U-TAE \cite{garnot2021panoptic}, the spatial feature from the last layer of the encoder is fused with the temporal attention. In 3D CNN \cite{Voelsen_Investigating2d3d_2022}, the fused feature is used after processing by the last encoder layer. For RNN-based models \cite{majd2020correlational}, the final hidden state is used. In TSVit \cite{Tarasiou_ViTsSITSVision_2023}, the used features are derived by extracting class token embeddings from the temporal transformer output.

\begin{figure}[t]
	\begin{center}
        \includegraphics[width=0.9\linewidth]{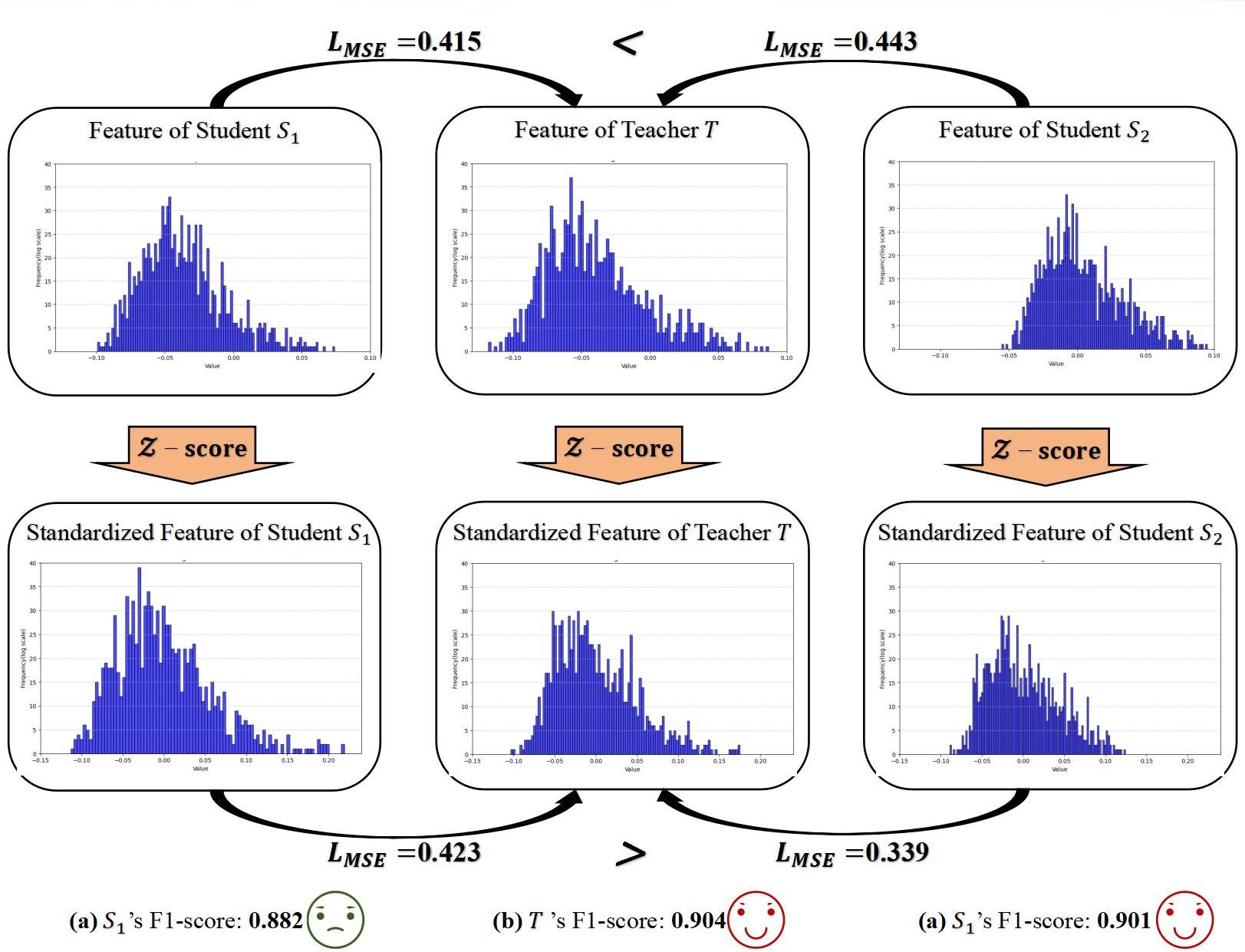}
		\caption{The toy case where two students, $S_1$ and $S_2$, are guided by the same teacher $T$. Before the standardization(Z-score), $S_1$  reconstructed features from incomplete SITS that are more similar to the teacher’s ideal features, yet it achieves a lower F1-score. In contrast, features from $S_2$ are less similar to ideal features and get a higher Mean Squared Error distance $L_{MSE}$, but they get a higher F1-score. After standardization, the issue is solved, which makes the model focus more on feature distribution instead of values and boundaries.}
		\label{F4}
	\end{center}
\end{figure}

Furthermore, we standardize the obtained features $F$ and $F^\prime$ to $Z$ and $Z^\prime$. As shown in Fig. \ref{F4}, this is because the temporal missing of samples can cause a global shift of the feature's distribution. Standardization helps the student model avoid paying excessive attention to the values and boundaries of the features, and instead pay more attention to the key patterns of features. Specifically, $F$ and $F^\prime$ are standardized along the channel dimensional, then the $f_S$ is trained by minimizing the squared $L2$ distance between $Z$ and $Z^\prime$:

\begin{equation}\label{equ2}
    Z = \frac{F - \mu_F}{\sigma_F}, \quad Z' = \frac{F' - \mu_{F'}}{\sigma_{F'}}, \quad
\end{equation}

\begin{equation}\label{equ3}
    \mathcal{L}_{\mathrm{FR}} = \frac{1}{N} \sum_{i=1}^{N} \left\| Z_i - Z_i' \right\|_2^2
\end{equation}
where $\mu_{\mathbf{F}}/\mu_{\mathbf{F}^\prime}$ and $\sigma_{\mathbf{F}}/\sigma_{\mathbf{F}^\prime}$ are the mean and standard deviation of $\mathbf{F}$ and $\mathbf{F}^\prime$ along the channel dimension.

Finally, in the prediction task, we not only expect the model to obtain the correct one-hot prediction, but also guide the model to mimic the teacher model's reasoning process. One the one hand, after inputting $X^\prime$, the prediction result $Y^\prime$ of $f_S$ are generated with support of $F^\prime$, which is supervised by ground truth label $Y$:

\begin{equation}\label{equ4}
    \mathcal{L}_{\mathrm{CE}} = -\frac{1}{N} \sum_{i=1}^{N} \sum_{k=1}^{C} Y_{i,k} \log\left( Y'_{i,k} \right)
\end{equation}

It also constrains the $f_S$ to reconstruct the critical patterns of $F^\prime$, which is beneficial for prediction. On the other hand, by inputting $X$ and $X^\prime$ to $f_T$ and $f_S$ respectively, we obtained logit outputs $l$ and $l^\prime$. The $f_S$ is trained by minimizing the Kullback-Leibler Divergence between the logits, thereby enabling the $f_S$ to learn how the $f_T$ handles the relative relationships between different classes:

\begin{equation}\label{equ5}
    \mathcal{L}_{\mathrm{KD}} = \mathrm{KLDivLoss} \left(
    \frac{\exp(l / T)}{\sum_{k=1}^{C} \exp(l^k / T)},\ 
    \frac{\exp(l' / T)}{\sum_{k=1}^{C} \exp(l'^k / T)}
    \right)
\end{equation}
where $T$ represents the distillation temperature, and $C$ represents the number of classes. The entire student model is jointly optimized by the overall loss $\mathcal{L}$, which is composed of $\mathcal{L}_{\mathrm{FR}}$ from the feature reconstruction task and $\mathcal{L}_{\mathrm{KD}}/\mathcal{L}_{\mathrm{CE}}$ from prediction tasks:
\begin{equation}\label{equ6}
    \mathcal{L} = \sigma \mathcal{L}_{\mathrm{FR}} + \gamma \mathcal{L}_{\mathrm{KD}} + \lambda \mathcal{L}_{\mathrm{CE}}
\end{equation}
where $\sigma$, $\gamma$, $\lambda$ are the weight coefficients for the respective losses. 

\section{Experiments and Results}
\label{sec4}

\subsection{Experiment settings}
\label{subsec4.1}

The experiments utilize the three datasets introduced in Section 3.1, Hunan SEN and Hunan PLA are used for cropland extraction, and the Fr\&Cat S4A Datasets are used for crop classification. Each $256 \times 256$ -pixel or $366 \times 366$ -pixel sample is normalized channel-wise by the mean and variance computed across all samples from the corresponding dataset. To validate the effectiveness and generalizability of the proposed method, all experiments were conducted in both simulated and real-world environments. For simulated experiments, we dropped 25\%-75\% time steps of each SITS sample from complete subsets, with a 60\%/40\% split for training and test sets, respectively. For real-world experiments, we employed the models trained in simulated environments, testing them on all samples from the incomplete subsets. All experiments use U-TAE \cite{garnot2021panoptic} as the backbone.

To evaluate the accuracy of our results, we employed multiple assessment metrics: Overall Accuracy (OA), Average Accuracy (AA), Mean Intersection over Union (mIoU), Kappa, and F1-scores for both individual classes and all classes (Mean F1-scores) collectively. The F1-score, serving as the harmonic mean of precision (User's Accuracy) and recall (Producer's Accuracy), effectively quantifies the model's classification reliability for cropland and crop types.

All models were trained using PyTorch on Ubuntu 16.04 operating system with an NVIDIA A6000 GPU (48 GB memory). Each model was trained using the AdamW optimizer, with a batch size of 4 and 100 epochs. The learning rate of our method was initially set to $1 \times 10^{-3}$

\subsection{Basic experiments}
\label{subsec4.2}

To validate the effectiveness of the proposed method for agricultural semantic segmentation tasks and its generalization capability across different sensors, we conducted experiments on cropland extraction and crop classification tasks using Sentinel-2 and PlanetScope SITS. All experiments were performed on three distinct datasets: Hunan SEN (based on Sentinel-2 SITS for cropland extraction), Hunan PLA (based on PlanetScope SITS for cropland extraction), and Fr\&Cat S4A (based on Sentinel-2 SITS for crop classification). As described in Section 3.1, each dataset is divided into complete and incomplete subsets to construct simulated and real-world experimental environments. 

\subsubsection{Compared methods}
\label{subsec4.2.1}

We compared proposed method with two method categories, namely Data Reconstruction (DR) and Data Augmentation (DA) for agricultural semantic segmentation tasks using SITS with time gaps. The Baseline method refers to training the model on 60\% of the SITS samples from the complete subsets and directly testing it on the simulated or real-world incomplete SITS without any adaptation.

\textbf{DR methods}: these methods perform comprehensive SITS reconstruction before model prediction. For direct interpolation, we implement three typical methods: Linear Interpolation (Linear-IN), Closest Interpolation ( Closest-IN), and Last Interpolation (Last-IN). For deep learning-based methods, we compare two state-of-the-art approaches: U-TILISE \cite{Stucker_UTILISESequencetoSequenceModel_2023} and GANFilling \cite{gonzalez2025generative}. These DR models were re-trained on the same data as our method, specifically using 60\% of complete subsets from each dataset. To simulate temporal missing scenarios, we generated sample pairs by applying 25\%-75\% temporal dropout to construct corresponding complete/incomplete SITS samples.

\textbf{DA methods}: these methods enhance prediction model robustness in simulated temporal missing scenarios by increasing training data diversity. We implement two typical methods in the temporal dimension: Window Slicing (DA-WS) \cite{Gao_DataAugmentationTimeSeries_2024} and Temporal Dropout (DA-TD) \cite{Yuan_EmpiricalStudyData_2025a}. WS randomly extracts continuous temporal windows covering 25\%-75\% of each SITS sample's full sequence length. TD randomly drops 25\%-75\% of time steps from each SITS sample. Each sample has a 50\% probability of being augmented during training.

\subsubsection{Cropland extraction with Sentinel-2 SITS}
\label{subsec4.2.2}

\begin{figure}[t]
	\begin{center}
        \includegraphics[width=1\linewidth]{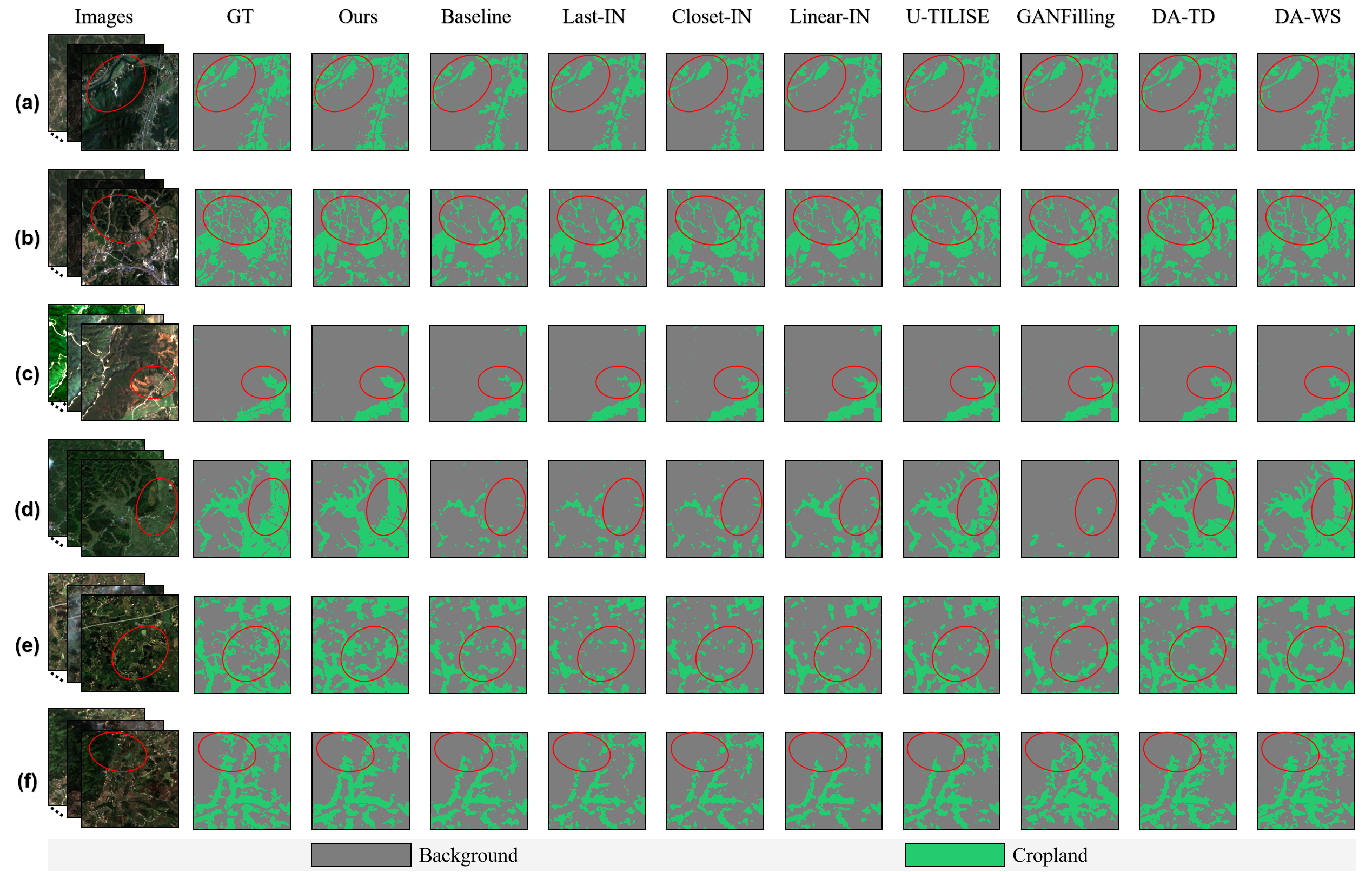}
		\caption{Visualization of sample results for our and compared methods in the Hunan SEN dataset: (a)-(c) from simulated experiments, (d)-(f) from real-world experiments.}
		\label{F5}
	\end{center}
\end{figure}

As shown in Table \ref{T1}, Table \ref{T2} ,and Fig. \ref{F5}, under the simulated and real-world environment, most of the data reconstruction (DR) methods failed; they even performed worse than the Baseline. Specifically, the cropland F1-score decreases by 0.93\% to 5.57\% compared to the Baseline in the simulated experiment. Except for U-TILISE, the cropland F1-score of other DR methods gdecreases by 6.87\% to 26.37\% compared to the baseline in real-world experiments. Due to insufficient information support, the inaccurately reconstructed images inevitably introduce noise into the entire SITS, which misleads the model and results in incorrect classification outcomes.

\begin{table}[htbp]
\renewcommand{\arraystretch}{1.3}
\centering
\resizebox{\textwidth}{!}{%
{\fontsize{9pt}{11pt}\selectfont
\begin{tabular}{llllllll}
\hline
Method     & OA(\%)  & AA(\%)  & Kappa & mIoU(\%) & M-F1(\%) & CL-F1(\%) & BG-F1(\%) \\ \hline
Baseline   & 89.53 & 87.87 & 0.71 & 75.15 & 85.26    & 77.34     & 93.19     \\
Last-IN    & 88.18 & 87.28 & 0.66 & 71.72 & 82.73    & 73.03     & 92.43     \\
Closest-IN  & 88.80 & 87.48 & 0.68 & 73.33 & 83.94    & 75.11     & 92.77     \\
Linear-IN  & 88.71 & 87.56 & 0.68 & 73.06 & 83.74    & 74.74     & 92.73     \\
U-TILIES   & 88.38 & 86.69 & 0.67 & 72.62 & 83.44    & 74.39     & 92.49     \\
GANFilling & 88.89 & 86.34 & 0.69 & 74.26 & 84.67    & 76.62     & 92.71     \\
DA-TD      & 89.57 & 87.88 & 0.71 & 75.25 & 85.34    & 77.47     & 93.21     \\
DA-WS      & 89.75 & 87.42 & 0.72 & 76.06 & 85.93    & 78.60     & 93.26     \\
Ours       & 90.84 & 88.12 & 0.75 & 78.66 & 87.72    & 81.53     & 93.91     \\ \hline
\end{tabular}
}
}
\caption{Performance of compared methods and our method in simulated experiments on the Hunan SEN dataset (CL-F1: cropland F1-score, BG-F1: background F1-score, M-F1: mean F1-scores).}
\label{T1}
\end{table}

\begin{table}[h!]
\renewcommand{\arraystretch}{1.3}
\centering
\resizebox{\textwidth}{!}{%
{\fontsize{9pt}{11pt}\selectfont
\begin{tabular}{llllllll}
\hline
Method     & OA(\%)  & AA(\%)  & Kappa & mIoU(\%) & M-F1(\%) & CL-F1(\%) & BG-F1(\%) \\ \hline
Baseline   & 85.78  & 83.31  & 0.41      & 57.83    & 69.46          & 47.14           & 91.78           \\
Last-IN    & 83.87  & 79.80  & 0.28      & 52.07    & 62.75          & 34.71           & 90.80           \\
Closest-IN  & 84.31  & 81.73  & 0.30      & 52.97    & 63.81          & 36.57           & 91.05           \\
Linear-IN  & 85.42  & 83.94  & 0.37      & 56.22    & 67.62          & 43.61           & 91.63           \\
U-TILIES   & 87.14  & 84.85  & 0.48      & 62.01    & 73.77          & 55.05           & 92.49           \\
GANFilling & 80.21  & 67.24  & 0.32      & 53.34    & 65.94          & 43.90           & 87.99           \\
DA-TD      & 88.72  & 87.38  & 0.56      & 66.37    & 77.80          & 62.23           & 93.37           \\
DA-WS      & 89.69  & 81.40  & 0.61      & 69.60    & 80.60          & 67.32           & 93.88           \\
Ours       & 89.49  & 83.98  & 0.64      & 71.14    & 81.99          & 70.37           & 93.61           \\ \hline
\end{tabular}}%
}
\caption{Performance of compared methods and our method in real-world experiments on the Hunan SEN dataset (CL-F1: cropland F1-score, BG-F1: background F1-score, M-F1: mean F1-scores).}
\label{T2}
\end{table}

DA methods outperform DR methods, compared to the Baseline, DA-TD, and DA-WS achieving improvements of 0.17\% and 1.63\%, respectively. By exposing the model to simulated temporal missing scenarios during training, these methods enable the model to learn how to match incomplete temporal features with correct labels under various temporal missing patterns. However, they are still limited by the model’s memory capacity, which hinders its ability to generalize across different missing patterns. Their generalization ability to unseen data is also affected by the shortcut learning phenomenon, where the model tends to rely on a jump connection for reasoning, rather than developing a robust reasoning capability that depends on long-term dependency. Moreover, the model may overfit on severely missing conditions (e.g., 50\%–75\% missing), and lose its ability to generalize to complete SITS or relatively mild missing conditions (e.g., 0\%–25\%). Detailed experimental results and analysis can be found in \ref{app1}. Unlike the similar performance of DA-WS and DA-TD in the simulated experiments, DA-WS shows an 8.18\% improvement over DA-TD in Cropland F1-score. This is due to the continuous-time gaps in real-world incomplete SITS, which are closer to the simulated training data of DA-WS. However, DA-WS does not show a significant advantage when applied to SITS collected from sensors with higher temporal resolution, and it exhibits substantial failure in crop classification tasks (details in section \ref{subsec4.2.3} and \ref{subsec4.2.4}).	

As shown in Table \ref{T1} and Table \ref{T2}, the results in the simulated and real-world experiments demonstrate that our method outperforms the other methods. Compared to the top-performing method, it shows an improvement of 4.53\% and 3.73\% in the cropland F1-score, and surpasses the baseline by 5.42\% and 49.28\% in cropland F1-score. The proposed method not only mitigates the negative impact of noise introduced during the comprehensive reconstruction process, but also leverages the reconstructed features to provide reliable support for the model to learn more robust reasoning capabilities.  Fig.5 shows the qualitative results of our methods and the compared methods in both simulated and real-world experiments. From Fig. \ref{F5} (a), (b), and (e), our method captures more fragmented and elongated croplands with smaller field sizes. In Fig. \ref{F5} (c), (d), and (f), it also demonstrates higher boundary completeness and retains more details.

Additionally, it can be observed that the baseline performs worse in real-world experiments compared to simulated experiments. This is also because the SITS in real-world experiments contain more continuous time gaps. The self-attention mechanism in U-TAE enables the model to dynamically allocate attentions, thereby reconstructing temporal dependencies by leveraging strongly correlated neighboring time steps. However, as the ratio of continuous time gaps increases, it becomes difficult for the model to reconstruct any information from the completely missing segments. A more detailed explanation can be found in Section \ref{subsec5.1}.

\subsubsection{Cropland extraction with PlanetScope SITS}
\label{subsec4.2.3}

Compared with Sentinel-2, the SITS from PlanetScope has a higher temporal resolution. This leads to a lower probability of continuous missing time steps in real-world SITS, with missing time steps tending to follow a more random distribution. This is because the availability of images increases during cloudy and rainy seasons (e.g., spring and winter in Hunan), while time gaps occur relatively more frequently in other seasons.

\begin{table}[htbp]
\renewcommand{\arraystretch}{1.3}
\centering
\resizebox{\textwidth}{!}{%
{\fontsize{9pt}{11pt}\selectfont
\begin{tabular}{llllllll}
\hline
Method     & OA(\%)  & AA(\%)  & Kappa & mIoU(\%) & M-F1(\%) & CL-F1(\%) & BG-F1(\%) \\ \hline
Baseline   & 87.87  & 87.22  & 0.67      & 72.34    & 83.30          & 74.56           & 92.04           \\
Last-IN    & 85.60  & 86.04  & 0.59      & 66.96    & 79.09          & 67.42           & 90.75           \\
Closest-IN  & 85.71  & 86.31  & 0.59      & 67.13    & 79.22          & 67.61           & 90.83           \\
Linear-IN  & 86.29  & 86.82  & 0.61      & 68.42    & 80.26          & 69.34           & 91.17           \\
U-TILIES   & 86.27  & 86.72  & 0.61      & 68.41    & 80.25          & 57.46           & 96.94           \\
GANFilling & 86.25  & 84.94  & 0.62      & 69.21    & 80.96          & 62.07           & 95.20           \\
DA-TD      & 88.94  & 86.50  & 0.71      & 75.57    & 85.67          & 78.84           & 92.51           \\
DA-WS      & 88.76  & 86.17  & 0.71      & 75.32    & 85.51          & 78.64           & 92.37           \\
Ours       & 89.97  & 88.14  & 0.74      & 77.40    & 86.90          & 80.57           & 93.24           \\ \hline
\end{tabular}}%
}
\caption{Performance of compared methods and our method in simulated experiments on the Hunan PLA dataset (CL-F1: cropland F1-score, BG-F1: background F1-score, M-F1: mean F1-scores).}
\label{T3}
\end{table}

\begin{table}[h!]
\renewcommand{\arraystretch}{1.3}
\centering
\resizebox{\textwidth}{!}{%
{\fontsize{9pt}{11pt}\selectfont
\begin{tabular}{llllllll}
\hline
Method     & OA(\%)  & AA(\%)  & Kappa & mIoU(\%) & M-F1(\%) & CL-F1(\%) & BG-F1(\%) \\ \hline
Baseline   & 90.60  & 88.53  & 0.72      & 76.27    & 85.97          & 77.90           & 94.03           \\
Last-IN    & 90.02  & 88.02  & 0.70      & 74.81    & 84.92          & 76.16           & 93.69           \\
Closest-IN  & 89.97  & 88.10  & 0.70      & 74.63    & 84.79          & 75.91           & 93.67           \\
Linear-IN  & 90.10  & 88.72  & 0.70      & 74.72    & 84.84          & 75.91           & 93.77           \\
U-TILIES   & 90.30  & 88.49  & 0.71      & 75.40    & 85.34          & 69.67           & 96.49           \\
GANFilling & 89.92  & 87.25  & 0.70      & 74.92    & 85.02          & 76.46           & 93.58           \\
DA-TD      & 90.41  & 86.08  & 0.74      & 77.30    & 86.77          & 79.82           & 93.71           \\
DA-WS      & 90.38  & 86.24  & 0.73      & 77.04    & 86.59          & 79.45           & 93.72           \\
Ours       & 90.67  & 88.34  & 0.76      & 79.17    & 88.10          & 82.58           & 93.63           \\ \hline
\end{tabular}}%
}
\caption{Performance of compared methods and our method in real-world experiments on the Hunan PLA dataset(CL-F1: cropland F1-score, BG-F1: background F1-score, M-F1: mean F1-scores).}
\label{T4}
\end{table}

\begin{figure}[t]
	\begin{center}
        \includegraphics[width=1\linewidth]{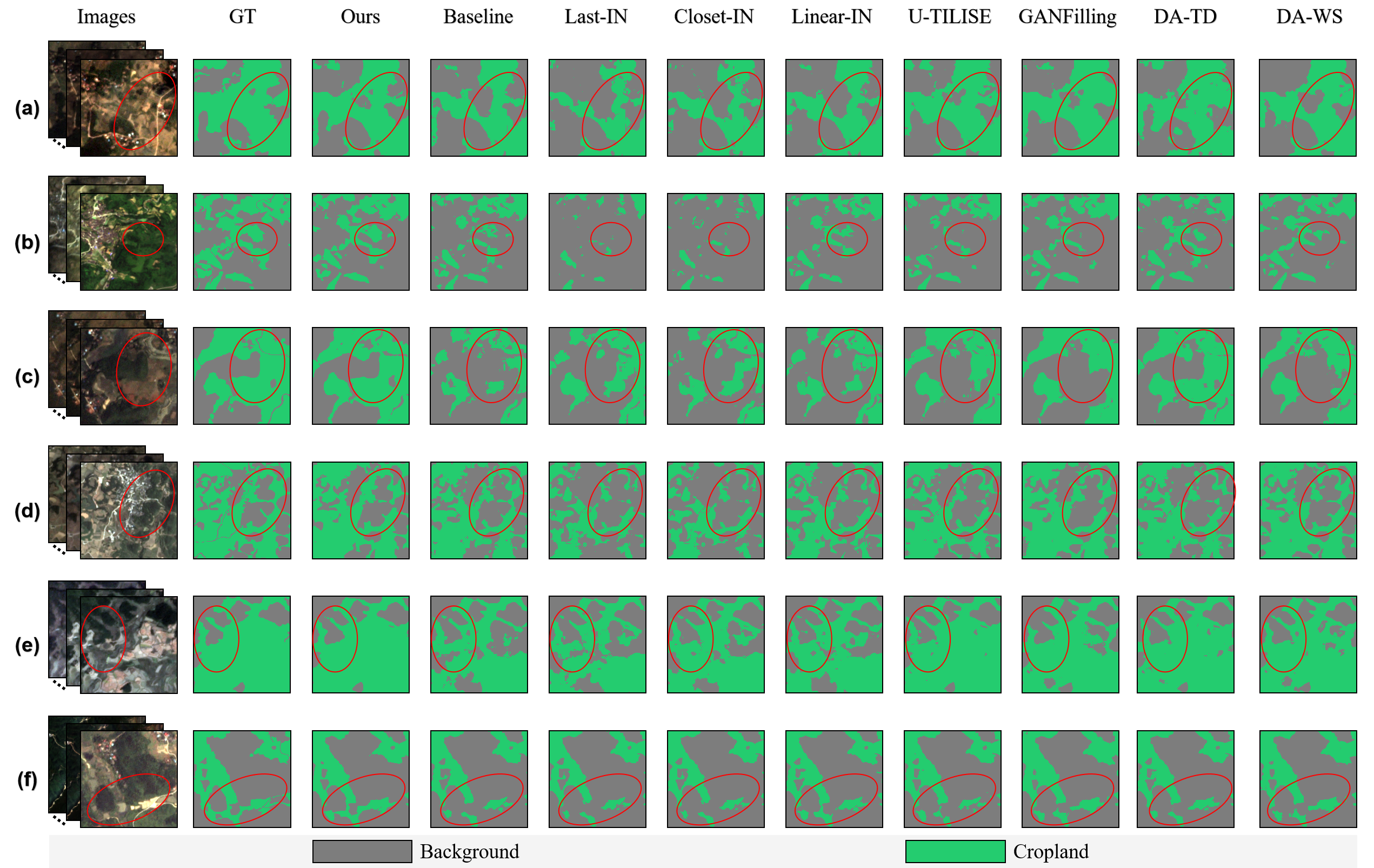}
		\caption{Visualization of sample results for our and compared methods in the Hunan PLA dataset: (a)-(c) from simulated experiments, (d)-(f) from real-world experiments.}
		\label{F6}
	\end{center}
\end{figure}

As shown in Table \ref{T3} and Table \ref{T4}, the performance trend of our method and the compared methods in the Hunan PLA dataset is similar to Hunan SEN. First, all DR methods still perform worse than the baseline, with cropland F1-score decreases ranging from 22.93\% to 7.00\% and 1.85\% to 10.56\% in the simulated and real-world experiments, respectively. Second, unlike the obvious advantage of DA-WS over DA-TD on the Hunan SEN dataset, DA-WS and DA-TD achieve similar results in real-world experiments, with cropland F1-scores of 79.82\% and 79.45\%, respectively. This demonstrates that the reduced probability of continuous missing time steps diminishes the advantage of DA-WS. Finally, our method also demonstrates strong robustness on SITS from PlanetScope in both simulated and real-world experiments. It surpassed the top-performing compared method by 2.19\% and 3.46\% in cropland F1-score, and outperformed the baseline by 8.06\% and 6.01\% in cropland F1-score. As shown in Fig. \ref{F6}, our method produces cropland extraction results with higher completeness and finer detail compared to other methods. 
For instance, in Fig. \ref{F6}(a) and (e), our method shows cropland boundaries that more closely resemble the labels. Fig. \ref{F6}(c) and (d) demonstrate more internal details within croplands as well as additional small cropland fields. Meanwhile, our method achieved a more accurate extraction of fields exhibiting unique phenological patterns that differ from other cropland fields within the same scene. For instance, in Fig. \ref{F6}(b) and (f), our method performs better on certain croplands with distinctive phenological patterns.

\subsubsection{Crop classification with Sentinel-2 SITS}
\label{subsec4.2.4}

As shown in Table \ref{T5} and Table \ref{T7}, compared with cropland extraction tasks, the performance of our method and the compared methods show different results in crop classification tasks. The DA methods no longer outperform DR methods, and most of the compared methods exhibit worse performance compared with the baseline. Specifically, in the simulated experiments, compared to the baseline, only DA-TD showed an increase in the mean F1-scores by 0.54\%, while the mean F1-scores of all other methods decreased by 1.37\% to 20.09\%. In the real-world experiments, all compared methods exhibited a decrease in mean F1-scores ranging from 6.05\% to 42.45\%. Firstly, limited by the model’s memory capability, the model trained with the DA-TD method is only able to adapt to the incomplete temporal features of certain categories, hindering their effectiveness in complex classification tasks.\begin{table}[t]
\renewcommand{\arraystretch}{1.3}
\centering
{\fontsize{9pt}{11pt}\selectfont
\begin{tabular}{llllll}
\hline
Method     & OA(\%) & AA(\%) & Kappa & mIoU(\%) & M-F1(\%) \\ \hline
Baseline   & 85.10  & 78.10  & 0.77  & 59.79    & 72.46    \\
Last-IN    & 84.01  & 84.26  & 0.76  & 57.36    & 69.96    \\
Closest-IN  & 84.76  & 86.29  & 0.77  & 59.02    & 71.47    \\
Linear-IN  & 84.38  & 82.00  & 0.76  & 56.65    & 69.45    \\
U-TILIES   & 81.87  & 72.66  & 0.73  & 48.42    & 60.93    \\
GANFilling & 78.62  & 72.20  & 0.67  & 43.50    & 57.90    \\
DA-TD      & 85.34  & 86.17  & 0.78  & 60.75    & 72.85    \\
DA-WS      & 80.49  & 60.26  & 0.70  & 45.34    & 55.32    \\
Ours       & 88.90  & 83.71  & 0.83  & 69.16    & 79.96   \\
\hline
\end{tabular}}
\caption{Performance of compared methods and our method in simulated experiments on the Fr\&Cat S4A dataset (M-F1: mean F1-scores).}
\label{T5}
\end{table}
\begin{table}[h!]
\renewcommand{\arraystretch}{1.3}
\centering
{\fontsize{9pt}{11pt}\selectfont
\begin{tabular}{lllllllll}
\hline
Method     & BC    & Cer   & V\&M  & OC    & R\&T  & LC    & SC    & OT    \\ \hline
Baseline   & 90.88 & 86.11 & 38.38 & 77.28 & 79.20 & 45.42 & 87.62 & 74.82 \\
Last-IN    & 90.43 & 84.67 & 31.08 & 79.40 & 73.65 & 39.96 & 87.89 & 72.60 \\
Closest-IN  & 90.67 & 85.79 & 34.95 & 80.50 & 74.44 & 41.86 & 89.43 & 74.15 \\
Linear-IN  & 90.60 & 85.17 & 31.92 & 79.81 & 67.64 & 40.19 & 86.26 & 73.98 \\
U-TILIES   & 90.02 & 80.57 & 12.92 & 65.77 & 52.32 & 29.36 & 85.05 & 71.40 \\
GANFilling & 87.48 & 77.95 & 23.18 & 60.43 & 53.58 & 33.37 & 61.40 & 65.83 \\
DA-TD      & 90.91 & 86.28 & 35.16 & 79.99 & 79.36 & 44.43 & 91.24 & 75.46 \\
DA-WS      & 87.44 & 84.59 & 0.00  & 0.00  & 73.50 & 42.98 & 83.99 & 70.02 \\
Ours       & 92.57 & 90.87 & 51.19 & 86.59 & 87.60 & 56.02 & 93.11 & 81.74 \\
\hline
\end{tabular}}
\caption{F1-scores(\%) of each class for compared and our method in simulated experiments on the Fr\&Cat S4A dataset, where BC, Cer, V\&M, OC, R\&T, LC, SC, and OT denote background, cereals, vegetables and melons, oilseed crops, root and tuber crops, leguminous crops, sugar crops, and other crops, respectively.}
\label{T6}
\end{table}
For example, as shown in Table \ref{T6} and Table \ref{T8}, in both simulated and real-world experiments, it achieves F1-scores of 86.28\%-91.24\% and 81.58\%-88.32\% on BC(Background), Cer(Cereals), and SC (Sugar crops), which are much higher than other categories. Secondly, the window slice might drop out whole growth cycles of certain crops, which makes the model trained under the DA-WS method tend to underfit these crops. For instance, in both simulated and real-world experiments, it achieves an F1-score of 0.00\% for V\&M (vegetables \& melons) and OC (oilseed crops). The growth cycle of V\&M is generally under 4 months, while the growth cycle of OC (oilseed crops), including soybeans, groundnuts, and other temporary oilseed crops, typically lasts no more than 5 months.

\begin{table}[t]
\renewcommand{\arraystretch}{1.3}
\centering
{\fontsize{9pt}{11pt}\selectfont
\begin{tabular}{llllll}
\hline
Method     & OA(\%) & AA(\%) & Kappa & mIoU(\%) & M-F1(\%) \\ \hline
Baseline   & 84.45  & 70.83  & 0.76  & 54.84    & 67.59    \\
Last-IN    & 82.78  & 79.77  & 0.74  & 50.85    & 62.48    \\
Closest-IN  & 82.85  & 80.54  & 0.74  & 51.80    & 63.50    \\
Linear-IN  & 82.87  & 76.94  & 0.74  & 48.03    & 59.66    \\
U-TILIES   & 79.62  & 49.88  & 0.69  & 33.82    & 44.14    \\
GANFilling & 82.29  & 67.45  & 0.73  & 48.12    & 60.61    \\
DA-TD      & 83.10  & 79.00  & 0.74  & 50.71    & 62.56    \\
DA-WS      & 76.10  & 41.73  & 0.62  & 30.01    & 38.90    \\
Ours       & 85.10  & 74.32  & 0.77  & 57.89    & 70.18   \\
\hline
\end{tabular}}
\caption{Performance of compared methods and our method in real-world experiments on the Fr\&Cat S4A dataset (M-F1: mean F1-scores).}
\label{T7}
\end{table}

\begin{table}[h!]
\renewcommand{\arraystretch}{1.3}
\centering
{\fontsize{9pt}{11pt}\selectfont
\begin{tabular}{lllllllll}
\hline
Method     & BC    & Cer   & V\&M  & OC    & R\&T  & LC    & SC    & OT    \\ \hline
Baseline   & 88.65 & 88.21 & 18.93 & 78.53 & 68.82 & 44.75 & 74.50 & 78.36 \\
Last-IN    & 88.11 & 83.06 & 6.06  & 78.23 & 59.60 & 25.50 & 83.03 & 76.23 \\
Closest-IN  & 88.13 & 83.17 & 7.85  & 78.25 & 65.33 & 25.87 & 82.98 & 76.38 \\
Linear-IN  & 88.21 & 82.79 & 9.16  & 78.22 & 40.14 & 23.01 & 78.98 & 76.77 \\
U-TILIES   & 87.99 & 79.09 & 8.86  & 51.99 & 11.12 & 20.67 & 19.48 & 73.91 \\
GANFilling & 87.94 & 83.62 & 18.14 & 77.17 & 23.67 & 44.58 & 74.74 & 75.03 \\
DA-TD      & 88.32 & 83.08 & 8.74  & 78.69 & 56.24 & 26.80 & 81.58 & 77.02 \\
DA-WS      & 83.90 & 84.26 & 0.00  & 0.00  & 15.47 & 32.84 & 26.64 & 68.07 \\
Ours       & 88.80 & 89.20 & 21.36 & 81.52 & 74.09 & 47.28 & 79.55 & 79.6 \\ 
\hline
\end{tabular}}
\caption{F1-scores(\%) of each class for compared and our method in real-world experiments on the Fr\&Cat S4A dataset, where BC, Cer, V\&M, OC, R\&T, LC, SC, and OT denote background, cereals, vegetables and melons, oilseed crops, root and tuber crops, leguminous crops, sugar crops, and other crops, respectively.}
\label{T8}
\end{table}

\begin{figure}[t]
	\begin{center}
        \includegraphics[width=1\linewidth]{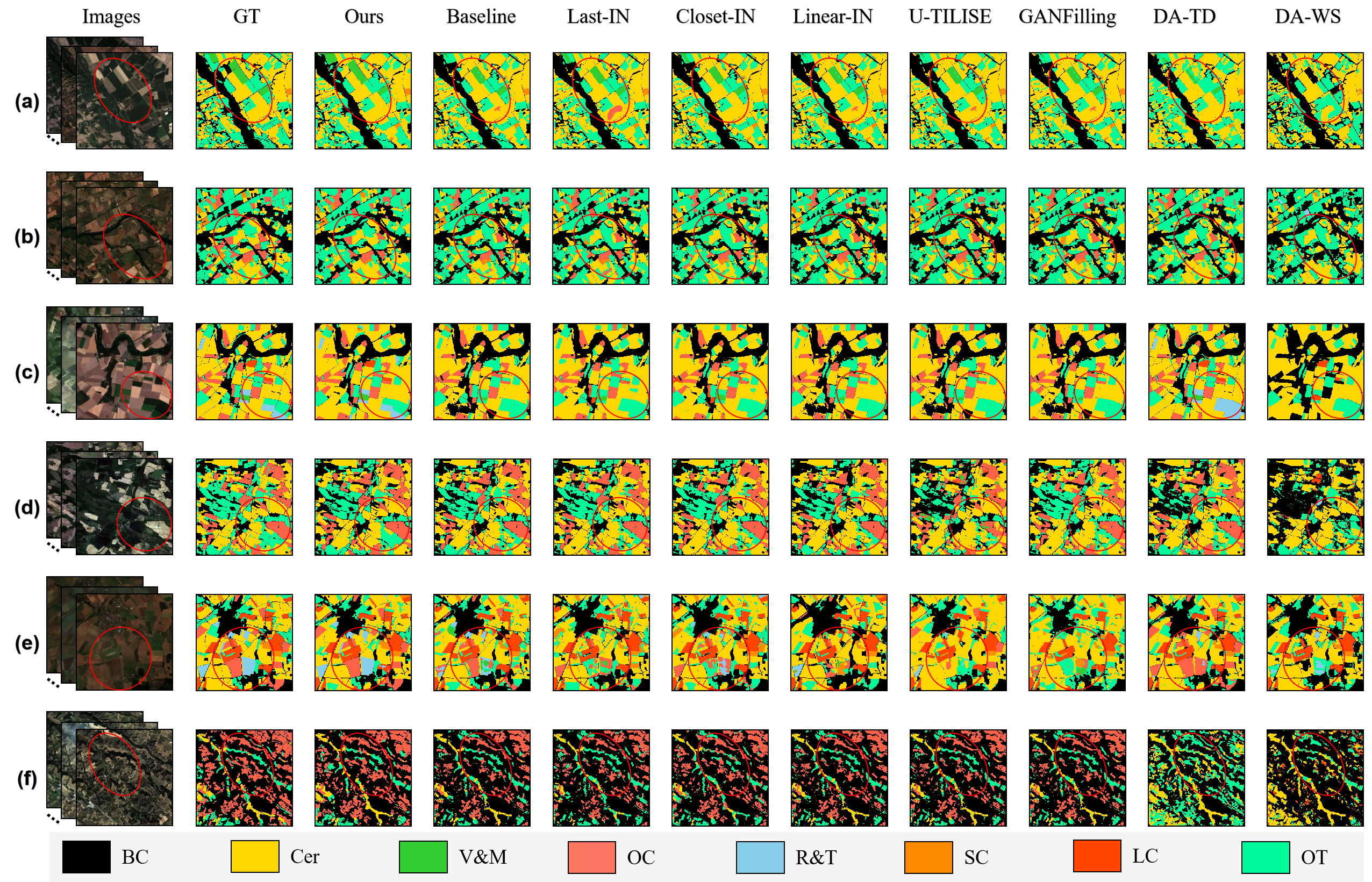}
		\caption{Visualization of sample results for our and compared methods in Fr\&Cat S4A dataset: (a)-(c) from simulated experiments, (d)-(f) from real-world experiments.}
		\label{F7}
	\end{center}
\end{figure}

Our method still outperforms all other compared methods. In simulated and real-world experiments, it achieved an improvement of 9.76\% and 10.52\% in mean F1-scores compared to the best-performing methods. It also achieves an improvement of 10.35\% and 3.83\% in the mean F1-scores compared to the baseline. The results demonstrate that our method also has strong generalization capability across different agricultural semantic segmentation tasks. As shown in Fig. \ref{F7}, in both simulated and real-world experiments, our method is able to accurately identify the boundaries between different crop fields and exhibits robustness in complex mixed-cropping scenarios. For instance, our method achieves more accurate boundary delineation for V\&M and Cer. 
In Fig. \ref{F7}(c) and (e), it successfully identifies multiple crop types in complex scenes, including OC, OT, SC, and Cer. In Fig. \ref{F7}(d) and (f), it provides more complete boundary extraction for fields with Cer and OT. 

However, our method also has its limitations. As shown in Table \ref{T6} and Table \ref{T8}, the V\&M(vegetables \& melons) and LC (leguminous crops) achieve lower accuracy compared to other crops, especially in real-world experimental environments, with only 21.36\% and 47.28\% in F1-score, respectively. For V\&M, the large and continuous time gaps also hinder our method from accurately reconstructing the fine temporal features of short-cycle crops. This leads to a lack of reliable references during the model reasoning process. As for LC, the high intra-class variability in their growth cycles limited the precise classification of backbones. The incorrect knowledge is transferred from the teacher model to the student model. For example, Broad Beans are typically planted in October and harvested in May, Chick Peas are planted in April and harvested in July, and Peas are planted in March and harvested in June.

\subsection{Ablation experiments}
\label{subsec4.3}

Although the last section has demonstrated the effectiveness of our proposed method, its robustness under varying levels of temporal incompleteness remains unclear. Therefore, we further investigate the effectiveness and limitations of our method by controlling the temporal missing rates in a simulated experiment environment. Specifically, we set the temporal missing rates from 0\% to 75\%. This allows us to evaluate the model’s performance on complete SITS and its robustness under low (25\%), middle (50\%), and high (75\%) levels of temporal incompleteness.

As shown in Table \ref{T9} and Table \ref{T10}, in cropland extraction tasks, our method maintains strong classification capability on complete time series and exhibits good generalization across different levels of temporal missing rates. Firstly, compared with the baseline on the complete SITS (0\%), our method achieves improvements of 1.67\% and 1.09\% in the cropland F1-score with the Hunan SEN and Hunan PLA datasets. It benefits from the model's feature reconstruction ability and self-distillation process [51]. The reconstructed features can support the model to avoid over-rely on shortcut reasoning paths. Knowledge distillation acts as a regularize function during the model's re-learning process by penalizing high variation in the local feature space, thereby mitigating the impact of overfitting \cite{Mobahi_SelfDistillationAmplifiesRegularization_2020}. Secondly, under low (25\%) and middle (50\%) levels of temporal missing conditions, our method achieves similar or increased performance compared with the baseline model’s performance on complete SITS (0\%). It maintains a cropland F1-score of over 81.26\% and 80.97\% on Hunan SEN and Hunan PLA, respectively. This indicates that our method can reconstruct key temporal features from limited temporal information, which supports the model to make accurate predictions. Thirdly, our method also faces challenges under a high level (75\%) of missing conditions, it only achieves cropland F1-scores of 79.36\% and 72.44\% on Hunan SEN and Hunan PLA, respectively. The relatively poorer performance on the Hunan PLA is due to the increased spatial resolution, which demands more refined reconstructed features for the model’s prediction. However, our method is more focused on reconstructing the global temporal features rather than restoring detailed information.

\begin{table}[htbp]
\renewcommand{\arraystretch}{1.3}
\centering
\resizebox{\textwidth}{!}{%
{\fontsize{9pt}{11pt}\selectfont
\begin{tabular}{llllllllll}
\hline
Mask ratio            & Methods  & OA(\%) & AA(\%) & Kappa & mIoU(\%) & M-F1(\%) & CL-F1(\%) & BG-F1(\%) \\ \hline
\multirow{2}{*}{0\%}  & Baseline & 90.69  & 88.53  & 0.75  & 78.06    & 87.29    & 80.72     & 93.86     \\
                      & Ours     & 91.14  & 88.64  & 0.76  & 79.24    & 88.09    & 82.07     & 94.12     \\ \hline
\multirow{2}{*}{25\%} & Baseline & 90.33  & 88.41  & 0.73  & 77.13    & 86.66    & 79.65     & 93.66     \\
                      & Ours     & 91.04  & 88.54  & 0.76  & 79.01    & 87.94    & 81.83     & 94.05     \\ \hline
\multirow{2}{*}{50\%} & Baseline & 89.47  & 87.87  & 0.70  & 75.03    & 85.18    & 77.20     & 93.15     \\
                      & Ours     & 90.80  & 88.32  & 0.75  & 78.47    & 87.58    & 81.26     & 93.90     \\ \hline
\multirow{2}{*}{75\%} & Baseline & 86.33  & 85.51  & 0.59  & 67.52    & 79.43    & 67.53     & 91.34     \\
                      & Ours     & 90.05  & 87.65  & 0.73  & 76.74    & 86.40    & 79.36     & 93.44    \\ 
\hline
\end{tabular}}%
}
\caption{Performance of the baseline and our method under different temporal missing rates on the Hunan SEN dataset. (CL-F1: cropland F1-score, BG-F1: background F1-score, M-F1: mean F1-scores)}
\label{T9}
\end{table}

\begin{table}[h!]
\renewcommand{\arraystretch}{1.3}
\centering
\resizebox{\textwidth}{!}{%
\begin{tabular}{llllllllll}
\hline
Mask ratio            & Methods  & OA(\%) & AA(\%) & Kappa & mIoU(\%) & M-F1(\%) & CL-F1(\%) & BG-F1(\%) \\ \hline
\multirow{2}{*}{0\%}  & Baseline & 90.47  & 88.66  & 0.75  & 78.48    & 87.63    & 81.69     & 93.56     \\
                      & Ours     & 90.67  & 88.34  & 0.76  & 79.17    & 88.10    & 82.58     & 93.63     \\ \hline
\multirow{2}{*}{25\%} & Baseline & 90.01  & 88.40  & 0.74  & 77.38    & 86.88    & 80.48     & 93.29     \\
                      & Ours     & 90.55  & 88.29  & 0.76  & 78.88    & 87.91    & 82.26     & 93.56     \\ \hline
\multirow{2}{*}{50\%} & Baseline & 87.85  & 87.37  & 0.67  & 72.21    & 83.19    & 74.35     & 92.04     \\
                      & Ours     & 90.13  & 88.25  & 0.74  & 77.76    & 87.15    & 80.97     & 93.33     \\ \hline
\multirow{2}{*}{75\%} & Baseline & 79.87  & 83.74  & 0.35  & 53.18    & 65.87    & 44.02     & 87.73     \\
                      & Ours     & 87.31  & 87.46  & 0.65  & 70.78    & 82.10    & 72.44     & 91.76    \\
\hline
\end{tabular}%
}
\caption{Performance of the baseline and our method under different temporal missing rates on the Hunan PLA dataset. (CL-F1: cropland F1-score, BG-F1: background F1-score, M-F1: mean F1-scores)}
\label{T10}
\end{table}

As shown in Table \ref{T11}, in the crop classification task, the performance of our method decreased with the increase of temporal missing rates. But it still keeps a mean F1-scores of 80.47\% under complete SITS, low (25\%), and middle (50\%) level missing conditions. According to the F1-score of each class presented in Table \ref{T12}, the overall decline is mainly attributed to the performance drop on V\&M and LC. When the temporal missing rate is 0\%-50\%, the F1-score for other classes only decreased from 0.19\% to 1.56\% compared to the baseline model’s performance on complete SITS (0\%). Specifically, the significant drop in V\&M is mainly because the increased temporal missing ratio may result complete absence of their growth cycles. Our method requires temporal anchor points to effectively recover temporal patterns. For LC, the decline is mainly due to the high intra-class variability, which prevents the teacher model from effectively learning discriminative features. Consequently, the transfer of incorrect knowledge during distillation may mislead the learning process of the student model. Under high-level missing conditions, our method still maintains a mean F1-scores of 70.70\%. Although it fails to recognize V\&M and LC, with F1-scores of only 35.01\% and 37.19\% respectively, it achieves F1-scores ranging from 73.74\% to 91.74\% for the remaining classes. In summary, the proposed method demonstrates limitations in the crop with short growth cycles and the limited by the teacher model’s classification ability. However, it can still reconstruct most of the key temporal features under varying degrees of temporal missing conditions in SITS, thereby supporting the model's reasoning process for most classes. 

\begin{table}[h!]
\renewcommand{\arraystretch}{1.3}
\centering
{\fontsize{9pt}{11pt}\selectfont
\begin{tabular}{lllllll}
\hline
Mask ratio            & Methods  & OA(\%) & AA(\%) & Kappa & mIoU(\%) & M-F1(\%) \\ \hline
\multirow{2}{*}{0\%}  & Baseline & 90.27  & 87.98  & 0.85  & 75.21    & 84.77    \\
                      & Ours     & 89.96  & 86.79  & 0.85  & 74.27    & 84.04    \\ \hline
\multirow{2}{*}{25\%} & Baseline & 89.32  & 86.10  & 0.84  & 71.63    & 82.05    \\
                      & Ours     & 89.75  & 86.36  & 0.85  & 73.23    & 83.19    \\ \hline
\multirow{2}{*}{50\%} & Baseline & 85.24  & 77.94  & 0.77  & 59.28    & 72.12    \\
                      & Ours     & 89.02  & 84.08  & 0.84  & 69.86    & 80.47    \\ \hline
\multirow{2}{*}{75\%} & Baseline & 72.89  & 61.59  & 0.57  & 35.17    & 48.50    \\
                      & Ours     & 85.91  & 76.14  & 0.79  & 58.21    & 70.70    \\ \hline
\end{tabular}
}
\caption{Performance of the baseline and our method under different temporal missing rates on the  Fr\&Cat S4A dataset. (M-F1: mean F1-scores)}
\label{T11}
\end{table}

\begin{table}[h!]
\renewcommand{\arraystretch}{1.3}
\centering
\resizebox{\textwidth}{!}{%
{\fontsize{9pt}{11pt}\selectfont
\begin{tabular}{llllllllll}
\hline
Mask ratio            & Methods  & BC    & Cer   & V\&M  & OC    & R\&T  & LC    & SC    & OT    \\ \hline
\multirow{2}{*}{0\%}  & Baseline & 93.03 & 92.46 & 66.23 & 88.55 & 93.93 & 64.89 & 95.51 & 83.61 \\ 
                      & Ours     & 92.85 & 92.27 & 62.97 & 88.31 & 93.21 & 64.25 & 95.15 & 83.34 \\ \hline
\multirow{2}{*}{25\%} & Baseline & 92.74 & 91.32 & 59.34 & 86.93 & 90.58 & 59.26 & 94.45 & 81.73 \\
                      & Ours     & 92.79 & 91.99 & 59.62 & 87.97 & 92.53 & 62.60 & 94.98 & 83.03 \\ \hline
\multirow{2}{*}{50\%} & Baseline & 91.27 & 86.32 & 42.16 & 77.25 & 76.45 & 42.08 & 86.65 & 74.80 \\
                      & Ours     & 92.60 & 90.93 & 51.84 & 86.63 & 89.17 & 56.87 & 93.80 & 81.93 \\ \hline
\multirow{2}{*}{75\%} & Baseline & 84.39 & 66.94 & 15.07 & 42.81 & 37.04 & 17.57 & 66.65 & 57.50 \\
                      & Ours     & 91.74 & 86.41 & 35.01 & 79.49 & 73.74 & 37.19 & 84.38 & 77.62 \\ \hline

\end{tabular}}%
}
\caption{F1-scores(\%) of the baseline and our method under different temporal missing rates on the Fr\&Cat S4A dataset (BC: background, Cer: cereals, V\&M: vegetables and melons, OC: oilseed crops, R\&T: root and tuber crops, LC: leguminous crops, SC: sugar crops, OT: other crops).}
\label{T12}
\end{table}

Additionally, the performance of DA methods is presented in \ref{app1} Although the model performs well in middle (50\%) and high (75\%) level missing conditions, it is prone to overfitting in these conditions, and loss its ability to handle complete SITS and low-level (25\%) missing conditions.

\section{Discussion}
\label{sec5}

\subsection{Networks agnostic of proposed framework}
\label{subsec5.1}

To evaluate the generalizability of the proposed framework across different network architectures in temporal missing conditions. We conducted experiments on typical types of network architectures: (1) Recurrent Neural Networks (RNN-based): Four backbones are used, including Convolutional Long Short-Term Memory (ConvLSTM) \cite{majd2020correlational}, Convolutional Gated Recurrent Unit (ConvGRU) \cite{siam2017convolutional}, U-Net with ConvLSTM (U-ConvLSTM)\cite{MRustowicz_SemanticSegmentationCrop_2019}, and Feature Pyramid Network with ConvLSTM (FPN-ConvLSTM)\cite{ChamorroMartinez_FullyConvolutionalRecurrent_2021a}. (2) 3D Convolutional Networks: We employ the 3D U-Net \cite{Voelsen_Investigating2d3d_2022}. (3) Self-attention-based Networks: Two backbones are utilized, including the Temporo-Spatial Vision Transformer (TSViT) \cite{Tarasiou_ViTsSITSVision_2023} and U-Net with Temporal Attention Encoder (U-TAE) \cite{garnot2021panoptic}. All experiments were conducted in the Hunan SEN dataset, involving both simulated and real-world scenarios on the complete and incomplete subsets. The baseline refers to the model trained in complete SITS samples from complete subsets and directly tested in simulated and real-world incomplete SITS. 

As shown in Table \ref{T13} and Table \ref{T14}, the proposed method shows improvement across all backbones in both simulated and real-world experiments. Specifically, in simulated experiments, our method maintains the Cropland F1-score above 80.00\% for most backbones. For RNN-based networks, it achieves improvements ranging from 1.78\% to 33.14\%. On the 3D U-Net, the Cropland F1-score increases by 47.81\%. For self-attention-based networks, improvements range from 4.83\% to 5.42\%. In real-world experiments, which involve more continuous time gaps, our method is still able to maintain the Cropland F1-score above 70.00\% for most backbones. On RNN-based networks, the Cropland F1-score improves by 2.00\% to 43.04\%. For the 3D U-Net, the improvement reaches 25.42\%, while for self-attention-based networks, it ranges from 29.06\% to 49.28\%. The network-agnostic ability of the proposed framework allows it to be flexibly adapted to a variety of real-world scenarios. It can be easily integrated into existing agricultural monitoring systems without requiring structural modifications, especially in environments with computational constraints or model-specific limitations.

\begin{table}[htbp]
\renewcommand{\arraystretch}{1.3}
\centering
\resizebox{\textwidth}{!}{%
\begin{tabular}{lllllllll}
\hline
Model                          & Method   & OA(\%) & AA(\%) & Kappa & mIoU(\%) & M-F1(\%) & CL-F1(\%) & BG-F1(\%) \\ \hline
\multirow{2}{*}{ConvLSTM}      & Baseline & 85.21  & 83.59  & 0.56  & 65.39    & 77.70    & 64.77     & 90.64     \\
                               & Ours     & 90.50  & 88.12  & 0.74  & 77.74    & 87.09    & 80.45     & 93.73     \\ \hline
\multirow{2}{*}{ConvGRU}       & Baseline & 86.70  & 86.30  & 0.60  & 68.16    & 79.93    & 68.28     & 91.58     \\
                               & Ours     & 90.48  & 88.29  & 0.74  & 77.60    & 86.99    & 80.24     & 93.73     \\ \hline
\multirow{2}{*}{UConvLSTM}     & Baseline & 90.36  & 88.18  & 0.74  & 77.29    & 86.77    & 79.89     & 93.66     \\
                               & Ours     & 90.80  & 88.28  & 0.75  & 78.50    & 87.60    & 81.31     & 93.90     \\ \hline
\multirow{2}{*}{FPN-ConvLSTIM} & Baseline & 81.08  & 74.99  & 0.48  & 60.76    & 74.17    & 60.81     & 87.53     \\
                               & Ours     & 90.68  & 88.22  & 0.75  & 78.19    & 87.39    & 80.96     & 93.83     \\ \hline
\multirow{2}{*}{3D U-Net}      & Baseline & 72.09  & 66.30  & 0.35  & 51.83    & 67.02    & 54.09     & 79.95     \\
                               & Ours     & 90.22  & 87.64  & 0.73  & 77.22    & 86.74    & 79.95     & 93.53     \\ \hline
\multirow{2}{*}{U-TAE}         & Baseline & 89.53  & 87.87  & 0.71  & 75.15    & 85.26    & 77.34     & 93.19     \\
                               & Ours     & 90.84 & 88.12 & 0.75 & 78.66 & 87.72    & 81.53     & 93.91      \\ \hline
\multirow{2}{*}{TSViT}         & Baseline & 89.31  & 87.42  & 0.70  & 74.66    & 84.92    & 76.77     & 93.06     \\
                               & Ours     & 90.45  & 87.91  & 0.74  & 77.72    & 87.08    & 80.48     & 93.68    \\ \hline
\end{tabular}}%

\caption{Evaluation of the proposed method under different networks in simulated experiments on the Hunan SEN dataset (CL-F1: cropland F1-score, BG-F1: background F1-score, M-F1: mean F1-scores).}
\label{T13}
\end{table}

\begin{table}[h!]
\renewcommand{\arraystretch}{1.3}
\centering
\resizebox{\textwidth}{!}{%
\begin{tabular}{lllllllll}
\hline
Model                          & Method   & OA(\%) & AA(\%) & Kappa & mIoU(\%) & M-F1(\%) & CL-F1(\%) & BG-F1(\%) \\ \hline
\multirow{2}{*}{ConvLSTM}      & Baseline & 84.07  & 74.63  & 0.53  & 63.80    & 76.45    & 63.06     & 89.84     \\ 
                               & Ours     & 89.23  & 82.61  & 0.65  & 71.57    & 82.41    & 71.46     & 93.36     \\ \hline
\multirow{2}{*}{ConvGRU}       & Baseline & 86.08  & 77.31  & 0.58  & 66.93    & 78.95    & 66.69     & 91.20     \\
                               & Ours     & 89.00  & 83.08  & 0.62  & 70.09    & 81.18    & 69.05     & 93.31     \\ \hline
\multirow{2}{*}{UConvLSTM}     & Baseline & 90.70  & 87.27  & 0.67  & 73.22    & 83.52    & 72.65     & 94.40     \\
                               & Ours     & 91.18  & 88.18  & 0.69  & 74.38    & 84.39    & 74.10     & 94.68     \\ \hline
\multirow{2}{*}{FPN-ConvLSTIM} & Baseline & 80.77  & 69.04  & 0.39  & 56.34    & 69.42    & 50.79     & 88.05     \\
                               & Ours     & 89.79  & 83.69  & 0.66  & 72.62    & 83.19    & 72.65     & 93.73     \\ \hline
\multirow{2}{*}{3D U-Net}      & Baseline & 84.31  & 74.06  & 0.48  & 61.48    & 74.02    & 57.68     & 90.37     \\
                               & Ours     & 90.44  & 85.48  & 0.67  & 72.88    & 83.29    & 72.35     & 94.22     \\ \hline
\multirow{2}{*}{U-TAE}         & Baseline & 85.78  & 83.31  & 0.41  & 57.83    & 69.46    & 47.14     & 91.78     \\
                               & Ours     & 89.49  & 83.98  & 0.64  & 71.14    & 81.99    & 70.37     & 93.61     \\ \hline
\multirow{2}{*}{TSViT}         & Baseline & 83.10  & 72.57  & 0.45  & 59.82    & 72.60    & 55.64     & 89.56     \\
                               & Ours     & 89.14  & 82.48  & 0.65  & 71.88    & 82.68    & 72.11     & 93.26    \\ \hline
\end{tabular}}%

\caption{Evaluation of the proposed method under different networks in real-world experiments on the Hunan SEN dataset (CL-F1: cropland F1-score, BG-F1: background F1-score, M-F1: mean F1-scores).}
\label{T14}
\end{table}

We further analyzed the distinct performance trends of different backbones in simulated and real-world experiments. The key findings are summarized as follows: (1) the baseline performance of 3D Convolutional Networks is relatively low in both simulated and real-world experiments. This is mainly because their spatiotemporal modeling is embedded in the most fundamental units of the network \cite{Tran_LearningSpatiotemporalFeatures_2015}. As a result, they struggle to properly align the temporal dimension when inputs are incomplete. This limitation not only restricts its ability to capture long-term temporal dependencies but also makes it difficult for the model to form a consistent understanding of diverse local temporal contexts. (2) In RNN-based networks, U-ConvLSTM performed much better than other backbones. Although these models all utilize recurrent units as their temporal processing module, U-ConvLSTM differs in its architectural design: it is fundamentally a U-Net, with ConvLSTM modules independently inserted into the skip connections between encoder and decoder layers \cite{MRustowicz_SemanticSegmentationCrop_2019}. This strategy reduces the sensitivity to temporal disruptions, thereby enabling the model to perform inference based on spatial information. In contrast, other backbones follow a classical recurrent architecture, where spatiotemporal information is propagated step by step. The missing time steps disrupt the hidden state propagation, making the model update its internal states based on incomplete or fragmented temporal context \cite{Sherstinsky_FundamentalsRecurrentNeural_2020}. As a result, errors accumulate over time, making the spatial information easily influenced by the temporal disturbance, which reduces the reliability of spatial features. (3) In simulated experiments, self-attention-based backbones show superior performance compared to most RNN-based backbones. The time gaps in simulated experiments are more random, leading to a higher frequency of temporal discontinuities. RNN-based architecture relies on sequential information propagation, where historical context is transferred step by step through hidden states. As the frequency of time gaps increases, these propagation pathways are repeatedly disrupted. The gating mechanisms are forced to perform unstable state updates frequently, which makes the network more easily exhibit gradient vanishing and explosion during the backpropagations \cite{Rehmer_VanishingExplodingGradient_2020}. Self-attention-based backbones rely on parallel dependency modeling, where temporal relationships are constructed through explicit interactions across all time steps. This allows the model to more easily reconstruct missing information caused by time gaps using strongly correlated observations \cite{Du_SAITSSelfattentionbasedImputation_2023}. (4) In contrast, self-attention-based backbones demonstrated inferior performance compared to RNN-based backbones in real-world experiments. As more continuous time gaps appear, the missing regions may lack any observations for interaction, making it extremely difficult to reconstruct information within those time gaps. Due to the strong correlation between adjacent time steps, the mechanism may concentrate attention on the time steps at both ends of the missing segment, which can completely fail the overall temporal modeling. In RNN-based backbones, the propagation of historical information is limited to disrupted, enabling the networks to preserve certain local temporal dependencies. It may support the networks to finish the prediction task.

\subsection{Impact of each module on proposed framework}
\label{subsec5.2}

To understand the individual contributions of each module within the proposed method, we conduct an ablation study by systematically removing or replacing specific components. Specifically, the proposed framework mainly consists of four modules: (1) Mask module (MA): Simulates temporal missing environments by producing incomplete SITS through random temporal masks, similar to the strategy used in DA-TD. (2) Distill module (KD): Employs a knowledge distillation framework, where soft supervision signals are provided by a teacher model trained on complete SITS.  (3) Feature reconstruction module (FR): Uses the temporal features extracted from complete SITS by the teacher model as supervised signals to guide the student model in learning feature reconstruction capabilities. (4) Feature standardization module (FS): Incorporates standardization steps within the FR Module to mitigate feature distribution shifts caused by time gaps, which prevent the model from overfitting to feature magnitudes. All experiments were conducted in the Hunan SEN dataset, involving both simulated and real-world scenarios on the complete and incomplete subsets. The U-TAE is used as the backbone.

As shown in Table \ref{T15}, in the simulated experiments, the model performance exhibits an upward trend as different modules are added. Specifically, the addition of the MA resulted in a slight improvement, with the Cropland F1 score increasing by 0.17\% compared to the baseline without any modules. As discussed in Sections 4.2 and \ref{app1}, the benefit of the MA module is partially diminished because simple data augmentation may lead the model to overfit under severe temporal missing conditions. With the addition of the KD module, the performance further improved by 1.67\% compared to the setting with only the MA module. This indicates that the soft supervision signal provides information about the relative probabilities of different classes, thereby enhancing the model’s generalization capability across varying scenarios. The FR module yielded an improvement of 1.16\% in Cropland F1-score over the MA+KD setting, while the subsequent inclusion of the FS module brought a larger gain of 3.52\%. It demonstrates that the standardization helps the model to pay more attention to the global distribution of features rather than local extreme values, thereby improving the model’s robustness under varied missing conditions. 

\begin{table}[htbp]
\renewcommand{\arraystretch}{1.3}
\centering
\resizebox{\textwidth}{!}{%
\begin{tabular}{lllllllllll}
\hline
\multicolumn{4}{c}{Module} & \multirow{2}{*}{OA(\%)} & \multirow{2}{*}{AA(\%)} & \multirow{2}{*}{Kappa} & \multirow{2}{*}{mIoU(\%)} & \multirow{2}{*}{M-F1(\%)} & \multirow{2}{*}{CL-F1(\%)} & \multirow{2}{*}{BG-F1(\%)} \\ \cline{1-4}
MA    & KD   & FR   & FS   &                         &                         &                        &                           &                           &                            &                            \\ \hline
×      & ×     & ×     & ×     & 89.53                   & 87.87                   & 0.71                   & 75.15                     & 85.26                     & 77.34                      & 93.19                      \\
\checkmark      & ×     & ×     & ×     & 89.57                   & 87.88                   & 0.71                   & 75.25                     & 85.34                     & 77.47                      & 93.21                      \\
\checkmark      & \checkmark     & ×     & ×     & 90.12                   & 88.59                   & 0.72                   & 76.43                     & 86.16                     & 78.76                      & 93.56                      \\
\checkmark      & \checkmark     & \checkmark     & ×     & 90.53                   & 89.14                   & 0.74                   & 77.29                     & 86.75                     & 79.67                      & 93.82                      \\
\checkmark      & \checkmark     & \checkmark     & \checkmark     & 90.84                   & 88.12                   & 0.75                   & 78.66                     & 87.72                     & 81.53                      & 93.91                      \\ \hline
\end{tabular}}%

\caption{Performance of Different Module Combinations in simulated experiments on the Hunan SEN dataset (CL-F1: cropland F1-score, BG-F1: background F1-score, M-F1: mean F1-scores).}
\label{T15}
\end{table}

As shown in Table \ref{T16}, in real-world experiments, the performance trend differs from that observed in the simulated experiments. Firstly, the addition of the MA module yielded a substantial improvement, increasing the Cropland F1 score by 32.01\%. As discussed in Section 5.1, this is because the robustness of the self-attention mechanism to temporal missingness becomes ineffective when facing continuous time gaps. In such cases, the benefits introduced by data augmentation far outweigh the potential negative effects caused by overfitting. Secondly, when only the FR module was added—without the FS module—it even resulted in a performance drop, with the Cropland F1 score decreasing by 3.04\%. This indicates that in real-world scenarios with more continuous time gaps, they may cause substantial distortions in feature magnitudes. Without standardization, forcing the model to mimic the complete temporal feature may lead the model to learn incorrect reconstruction abilities. The addition of the FS module led to a 13.21\% improvement in Cropland F1-score compared to the setting with the other three modules. This further demonstrates the necessity of feature standardization. Additional experiments and analyses under different missing conditions are provided in \ref{app2}.

\begin{table}[htbp]
\renewcommand{\arraystretch}{1.3}
\centering
\resizebox{\textwidth}{!}{%
\begin{tabular}{lllllllllll}
\hline
\multicolumn{4}{c}{Module} & \multirow{2}{*}{OA(\%)} & \multirow{2}{*}{AA(\%)} & \multirow{2}{*}{Kappa} & \multirow{2}{*}{mIoU(\%)} & \multirow{2}{*}{M-F1(\%)} & \multirow{2}{*}{CL-F1(\%)} & \multirow{2}{*}{BG-F1(\%)} \\ \cline{1-4}
MA    & KD   & FR   & FS   &                         &                         &                        &                           &                           &                            &                            \\ \hline
×     & ×    & ×    & ×    & 85.78                   & 83.31                   & 0.41                   & 57.83                     & 69.46                     & 47.14                      & 91.78                      \\
\checkmark     & ×    & ×    & ×    & 88.72                   & 87.38                   & 0.56                   & 66.37                     & 77.80                     & 62.23                      & 93.37                      \\
\checkmark     & \checkmark    & ×    & ×    & 88.60                   & 84.87                   & 0.58                   & 67.25                     & 78.67                     & 64.11                      & 93.22                      \\
\checkmark     & \checkmark    & \checkmark    & ×    & 88.83                   & 88.21                   & 0.56                   & 66.40                     & 77.81                     & 62.16                      & 93.45                      \\
\checkmark     & \checkmark    & \checkmark    & \checkmark    & 89.49                   & 83.98                   & 0.64                   & 71.14                     & 81.99                     & 70.37                      & 93.61                      \\ \hline

\end{tabular}}%

\caption{Performance of Different Module Combinations in real-world experiments on the Hunan SEN dataset (CL-F1: cropland F1-score, BG-F1: background F1-score, M-F1: mean F1-scores).}
\label{T16}
\end{table}

\section{Conclusion}
\label{sec6}

In this paper, we propose a feature reconstruction and prediction joint learning framework for multi-temporal agricultural segmentation tasks. The framework is designed to enhance the model's robustness under scenarios with missing temporal observations. It can selectively reconstruct key temporal features, thereby enabling the model to obtain reasoning ability on incomplete SITS. Experimental results demonstrate that the proposed framework achieves significant improvements over other state-of-the-art methods across three study areas. More importantly, it exhibits strong generalization capabilities in multiple aspects: (1) generalization across different agricultural semantic segmentation tasks, including cropland extraction and crop classification; (2) generalization across different sensors, primarily Sentinel-2 and PlanetScope; (3) generalization under different numbers of time gaps, while retaining classification performance on complete SITS; and (4) agnostic across different model architectures, including seven typical backbones. This greatly enhances the applicability of the proposed method, enabling it to be efficiently employed in diverse agricultural monitoring scenarios. In addition, we explored and analyzed the contributions of different modules within the proposed method, which enhanced its interpretability. Future research will focus on further investigating the model's behavior under various temporal gap patterns and on reconstructing more fine-grained temporal features to support more complex tasks.

\section*{Acknowledgements}
This work was supported in part by the Natural Science Foundation of Hunan for Distinguished Young Scholars under Grant 2022JJ10072; in part by the National Natural Science Foundation of China under Grant 42471419 and Grant 42171376.

\appendix
\section{Performance of DA Methods Under Varying Temporal Missing Conditions}
\label{app1}

The experimental setup is the same as Section \ref{subsec4.3}, where simulated experiments are conducted on different datasets under varying temporal missing rates. The missing rate is set from 0\% to 25\%. As shown in Table \ref{T17} and Table \ref{T18}, in the cropland extraction task, DA methods exhibit similar or even decreased performance compared to the Baseline under complete SITS (0\%) and low missing rate (25\%) conditions. While DA methods achieve better performance than the Baseline under middle (50\%) and high (75\%) missing rates, they still underperform compared to our method. This demonstrates that the DA methods have overfitted to the middle (50\%) and high (75\%) missing conditions, and lose their ability to handle complete SITS (0\%) and low (25\%) missing conditions. In contrast, with the support of reconstructed features, the model achieves better performance in middle (50\%) and high (75\%) missing conditions compared with DA methods. It also optimized the model's ability to handle complete SITS and keep its robustness under low missing rate (25\%) conditions.

\begin{table}[htbp]
\renewcommand{\arraystretch}{1.3}
\centering
\resizebox{\textwidth}{!}{%
\begin{tabular}{lllllllll}
\hline
Mask ratio             & Method & OA(\%)                                               & AA(\%)                                               & Kappa                                                & mIoU(\%)                                             & M-F1(\%)                                             & CL-F1(\%)                                            & BG-F1(\%)                                            \\ \hline
                       & DA-TD  & \cellcolor[HTML]{FFC7CE}{\color[HTML]{9C0006} -0.21} & 0.16                                                 & \cellcolor[HTML]{FFC7CE}{\color[HTML]{9C0006} -0.01} & \cellcolor[HTML]{FFC7CE}{\color[HTML]{9C0006} -0.78} & \cellcolor[HTML]{FFC7CE}{\color[HTML]{9C0006} -0.48} & \cellcolor[HTML]{FFC7CE}{\color[HTML]{9C0006} -0.92} & \cellcolor[HTML]{FFC7CE}{\color[HTML]{9C0006} -0.10} \\
                       & DA-WS  & \cellcolor[HTML]{FFC7CE}{\color[HTML]{9C0006} -0.60} & \cellcolor[HTML]{FFC7CE}{\color[HTML]{9C0006} -0.81} & \cellcolor[HTML]{FFC7CE}{\color[HTML]{9C0006} -0.02} & \cellcolor[HTML]{FFC7CE}{\color[HTML]{9C0006} -1.45} & \cellcolor[HTML]{FFC7CE}{\color[HTML]{9C0006} -0.87} & \cellcolor[HTML]{FFC7CE}{\color[HTML]{9C0006} -1.45} & \cellcolor[HTML]{FFC7CE}{\color[HTML]{9C0006} -0.37} \\ 
\multirow{-3}{*}{0\%}  & Ours   & 0.50                                                 & 0.12                                                 & 0.02                                                 & 1.51                                                 & 0.92                                                 & 1.67                                                 & 0.28                                                 \\ \hline
                       & DA-TD  & 0.09                                                 & 0.27                                                 & 0.00                                                 & 0.12                                                 & 0.07                                                 & 0.08                                                 & 0.06                                                 \\ 
                       & DA-WS  & \cellcolor[HTML]{FFC7CE}{\color[HTML]{9C0006} -0.37} & \cellcolor[HTML]{FFC7CE}{\color[HTML]{9C0006} -0.88} & \cellcolor[HTML]{FFC7CE}{\color[HTML]{9C0006} -0.01} & \cellcolor[HTML]{FFC7CE}{\color[HTML]{9C0006} -0.67} & \cellcolor[HTML]{FFC7CE}{\color[HTML]{9C0006} -0.40} & \cellcolor[HTML]{FFC7CE}{\color[HTML]{9C0006} -0.55} & \cellcolor[HTML]{FFC7CE}{\color[HTML]{9C0006} -0.27} \\ 
\multirow{-3}{*}{25\%} & Ours   & 0.79                                                 & 0.15                                                 & 0.03                                                 & 2.44                                                 & 1.48                                                 & 2.74                                                 & 0.42                                                 \\ \hline
                       & DA-TD  & 0.74                                                 & 0.81                                                 & 0.03                                                 & 1.96                                                 & 1.21                                                 & 2.15                                                 & 0.45                                                 \\
                       & DA-WS  & 0.26                                                 & \cellcolor[HTML]{FFC7CE}{\color[HTML]{9C0006} -0.60} & 0.02                                                 & 1.20                                                 & 0.77                                                 & 1.63                                                 & 0.09                                                 \\ 
\multirow{-3}{*}{50\%} & Ours   & 1.49                                                 & 0.51                                                 & 0.07                                                 & 4.58                                                 & 2.82                                                 & 5.26                                                 & 0.81                                                 \\ \hline
                       & DA-TD  & 3.16                                                 & 3.15                                                 & 0.16                                                 & 9.23                                                 & 6.04                                                 & 11.82                                                & 1.77                                                 \\
                       & DA-WS  & 2.65                                                 & 1.09                                                 & 0.15                                                 & 8.74                                                 & 5.80                                                 & 11.80                                                & 1.37                                                 \\ 
\multirow{-3}{*}{75\%} & Ours   & 4.31                                                 & 2.50                                                 & 0.23                                                 & 13.66                                                & 8.78                                                 & 17.52                                                & 2.30                                                
\\ \hline
\end{tabular}}%
\caption{The relative increase or decrease in performance of our and DA methods compared with baseline methods under different temporal missing rates on the Hunan SEN dataset. (CL-F1: cropland F1-score, BG-F1: background F1-score, M-F1: mean F1-scores). Values in red indicate a decrease compared to the Baseline.}
\label{T17}
\end{table}

\begin{table}[htbp]
\renewcommand{\arraystretch}{1.3}
\centering
\resizebox{\textwidth}{!}{%
\begin{tabular}{lllllllll}
\hline
Mask ratio             & Method & OA(\%)                                               & AA(\%)                                               & Kappa                                                & mIoU(\%)                                             & M-F1(\%)                                             & CL-F1(\%)                                            & BG-F1(\%)                                            \\ \hline
                       & DA-TD  & \cellcolor[HTML]{FFC7CE}{\color[HTML]{9C0006} -1.50} & \cellcolor[HTML]{FFC7CE}{\color[HTML]{9C0006} -3.10} & \cellcolor[HTML]{FFC7CE}{\color[HTML]{9C0006} -0.03} & \cellcolor[HTML]{FFC7CE}{\color[HTML]{9C0006} -2.45} & \cellcolor[HTML]{FFC7CE}{\color[HTML]{9C0006} -1.40} & \cellcolor[HTML]{FFC7CE}{\color[HTML]{9C0006} 0.62}  & \cellcolor[HTML]{FFC7CE}{\color[HTML]{9C0006} -2.02} \\
                       & DA-WS  & \cellcolor[HTML]{FFC7CE}{\color[HTML]{9C0006} -1.29} & \cellcolor[HTML]{FFC7CE}{\color[HTML]{9C0006} -2.77} & \cellcolor[HTML]{FFC7CE}{\color[HTML]{9C0006} -0.03} & \cellcolor[HTML]{FFC7CE}{\color[HTML]{9C0006} -2.09} & \cellcolor[HTML]{FFC7CE}{\color[HTML]{9C0006} -1.20} & \cellcolor[HTML]{FFC7CE}{\color[HTML]{9C0006} -1.41} & \cellcolor[HTML]{FFC7CE}{\color[HTML]{9C0006} -1.00} \\
\multirow{-3}{*}{0\%}  & Ours   & 0.22                                                 & \cellcolor[HTML]{FFC7CE}{\color[HTML]{9C0006} -0.36} & 0.01                                                 & 0.88                                                 & 0.54                                                 & 1.09                                                 & 0.07                                                 \\ \hline
                       & DA-TD  & \cellcolor[HTML]{FFC7CE}{\color[HTML]{9C0006} -0.86} & \cellcolor[HTML]{FFC7CE}{\color[HTML]{9C0006} -2.43} & \cellcolor[HTML]{FFC7CE}{\color[HTML]{9C0006} -0.01} & \cellcolor[HTML]{FFC7CE}{\color[HTML]{9C0006} -0.98} & \cellcolor[HTML]{FFC7CE}{\color[HTML]{9C0006} -0.52} & \cellcolor[HTML]{FFC7CE}{\color[HTML]{9C0006} -0.30} & \cellcolor[HTML]{FFC7CE}{\color[HTML]{9C0006} -0.73} \\
                       & DA-WS  & \cellcolor[HTML]{FFC7CE}{\color[HTML]{9C0006} -0.87} & \cellcolor[HTML]{FFC7CE}{\color[HTML]{9C0006} -2.45} & \cellcolor[HTML]{FFC7CE}{\color[HTML]{9C0006} -0.01} & \cellcolor[HTML]{FFC7CE}{\color[HTML]{9C0006} -0.98} & \cellcolor[HTML]{FFC7CE}{\color[HTML]{9C0006} -0.52} & \cellcolor[HTML]{FFC7CE}{\color[HTML]{9C0006} -0.27} & \cellcolor[HTML]{FFC7CE}{\color[HTML]{9C0006} -0.74} \\ 
\multirow{-3}{*}{25\%} & Ours   & 0.60                                                 & \cellcolor[HTML]{FFC7CE}{\color[HTML]{9C0006} -0.12} & 0.03                                                 & 1.94                                                 & 1.19                                                 & 2.21                                                 & 0.29                                                 \\ \hline
                       & DA-TD  & 1.46                                                 & \cellcolor[HTML]{FFC7CE}{\color[HTML]{9C0006} -0.77} & 0.08                                                 & 5.19                                                 & 3.32                                                 & 6.60                                                 & 0.64                                                 \\
                       & DA-WS  & 1.13                                                 & \cellcolor[HTML]{FFC7CE}{\color[HTML]{9C0006} -1.27} & 0.07                                                 & 4.50                                                 & 2.91                                                 & 5.97                                                 & 0.42                                                 \\
\multirow{-3}{*}{50\%} & Ours   & 2.60                                                 & 1.01                                                 & 0.12                                                 & 7.69                                                 & 4.76                                                 & 8.90                                                 & 1.40                                                 \\ \hline
                       & DA-TD  & 8.11                                                 & 3.13                                                 & 0.74                                                 & 29.30                                                & 22.27                                                & 58.81                                                & 3.92                                                 \\
                       & DA-WS  & 7.42                                                 & 2.01                                                 & 0.70                                                 & 27.45                                                & 21.13                                                & 56.20                                                & 3.51                                                 \\
\multirow{-3}{*}{75\%} & Ours   & 9.32                                                 & 4.44                                                 & 0.83       &33.10 	&24.64 	&64.56 	&4.59 
      
\\ \hline

\end{tabular}}%
\caption{The relative increase or decrease in performance of our and DA methods compared with baseline methods under different temporal missing rates on the Hunan PLA dataset. (CL-F1: cropland F1-score, BG-F1: background F1-score, M-F1: mean F1-scores). Values in red indicate a decrease compared to the Baseline.}
\label{T18}
\end{table}

As shown in Table \ref{T19}, in the crop classification task, unlike the DA methods, which suffer a significant performance decrease on complete SITS, our method shows only a 0.86\% drop in mean F1-scores. indicating minimal impact on the model’s ability to handle complete temporal inputs. Under the 25\% missing rate, our method is the only one that achieves performance improvement over the Baseline, whereas DA-based methods show declines across nearly all metrics. At 50\% and 75\% missing rates, although the DA-TD method shows improvements in mean F1-scores (1.94\% and 26.31\%, respectively), it is still significantly lower than the improvements achieved by our method under the same conditions (11.58\% and 45.77\%, respectively).

\begin{table}[htbp]
\renewcommand{\arraystretch}{1.3}
\centering
{\fontsize{9pt}{11pt}\selectfont
\begin{tabular}{lllllllll}
\hline
Mask ratio             & Method & OA(\%)                                               & AA(\%)                                                & Kappa                                                & mIoU(\%)                                              & M-F1(\%)                                              \\ 
                       & DA-TD  & \cellcolor[HTML]{FFC7CE}{\color[HTML]{9C0006} -4.04} & 1.47                                                  & \cellcolor[HTML]{FFC7CE}{\color[HTML]{9C0006} -0.06} & \cellcolor[HTML]{FFC7CE}{\color[HTML]{9C0006} -11.20} & \cellcolor[HTML]{FFC7CE}{\color[HTML]{9C0006} -7.92}  \\
                       & DA-WS  & \cellcolor[HTML]{FFC7CE}{\color[HTML]{9C0006} -8.98} & \cellcolor[HTML]{FFC7CE}{\color[HTML]{9C0006} -28.12} & \cellcolor[HTML]{FFC7CE}{\color[HTML]{9C0006} -0.15} & \cellcolor[HTML]{FFC7CE}{\color[HTML]{9C0006} -31.72} & \cellcolor[HTML]{FFC7CE}{\color[HTML]{9C0006} -29.09} \\
\multirow{-3}{*}{0\%}  & Ours   & \cellcolor[HTML]{FFC7CE}{\color[HTML]{9C0006} -0.34} & \cellcolor[HTML]{FFC7CE}{\color[HTML]{9C0006} -1.35}  & \cellcolor[HTML]{FFC7CE}{\color[HTML]{9C0006} 0.00}  & \cellcolor[HTML]{FFC7CE}{\color[HTML]{9C0006} -1.25}  & \cellcolor[HTML]{FFC7CE}{\color[HTML]{9C0006} -0.86}  \\ \hline
                       & DA-TD  & \cellcolor[HTML]{FFC7CE}{\color[HTML]{9C0006} -3.28} & 2.83                                                  & \cellcolor[HTML]{FFC7CE}{\color[HTML]{9C0006} -0.05} & \cellcolor[HTML]{FFC7CE}{\color[HTML]{9C0006} -8.61}  & \cellcolor[HTML]{FFC7CE}{\color[HTML]{9C0006} -6.20}  \\
                       & DA-WS  & \cellcolor[HTML]{FFC7CE}{\color[HTML]{9C0006} -8.40} & \cellcolor[HTML]{FFC7CE}{\color[HTML]{9C0006} -27.46} & \cellcolor[HTML]{FFC7CE}{\color[HTML]{9C0006} -0.14} & \cellcolor[HTML]{FFC7CE}{\color[HTML]{9C0006} -30.66} & \cellcolor[HTML]{FFC7CE}{\color[HTML]{9C0006} -28.36} \\
\multirow{-3}{*}{25\%} & Ours   & 0.48                                                 & 0.30                                                  & 0.01                                                 & 2.23                                                  & 1.39                                                  \\  \hline
                       & DA-TD  & 0.36                                                 & 11.20                                                 & 0.01                                                 & 3.71                                                  & 1.94                                                  \\
                       & DA-WS  & \cellcolor[HTML]{FFC7CE}{\color[HTML]{9C0006} -5.43} & \cellcolor[HTML]{FFC7CE}{\color[HTML]{9C0006} -22.85} & \cellcolor[HTML]{FFC7CE}{\color[HTML]{9C0006} -0.10} & \cellcolor[HTML]{FFC7CE}{\color[HTML]{9C0006} -24.16} & \cellcolor[HTML]{FFC7CE}{\color[HTML]{9C0006} -24.02} \\
\multirow{-3}{*}{50\%} & Ours   & 4.43                                                 & 7.88                                                  & 0.08                                                 & 17.85                                                 & 11.58                                                 \\ \hline
                       & DA-TD  & 11.39                                                & 27.08                                                 & 0.26                                                 & 36.88                                                 & 26.31                                                 \\
                       & DA-WS  & 4.28                                                 & \cellcolor[HTML]{FFC7CE}{\color[HTML]{9C0006} -11.25} & 0.10                                                 & \cellcolor[HTML]{FFC7CE}{\color[HTML]{9C0006} -2.13}  & \cellcolor[HTML]{FFC7CE}{\color[HTML]{9C0006} -7.90}  \\ 
\multirow{-3}{*}{75\%} & Ours   & 17.86                                                & 23.62                                                 & 0.39                                                 & 65.51                                                 & 45.77                                                \\   \hline

\end{tabular}}
\caption{The relative increase or decrease in performance of our and DA methods compared with baseline methods under different temporal missing rates on the Fr\&Cat S4A dataset. (M-F1: mean F1-scores). Values in red indicate a decrease compared to the Baseline.}
\label{T19}
\end{table}

\section{Effect of different module combinations under varying temporal missing conditions}
\label{app2}

Our exploration of each module remains relatively limited. Therefore, we further investigate the effectiveness of modules under varying levels of temporal missingness. In addition, we conduct similar experiments on the crop classification task. The experiment settings are the same as section \ref{subsec4.3}. 

\begin{figure}[t]
	\begin{center}
        \includegraphics[width=1\linewidth]{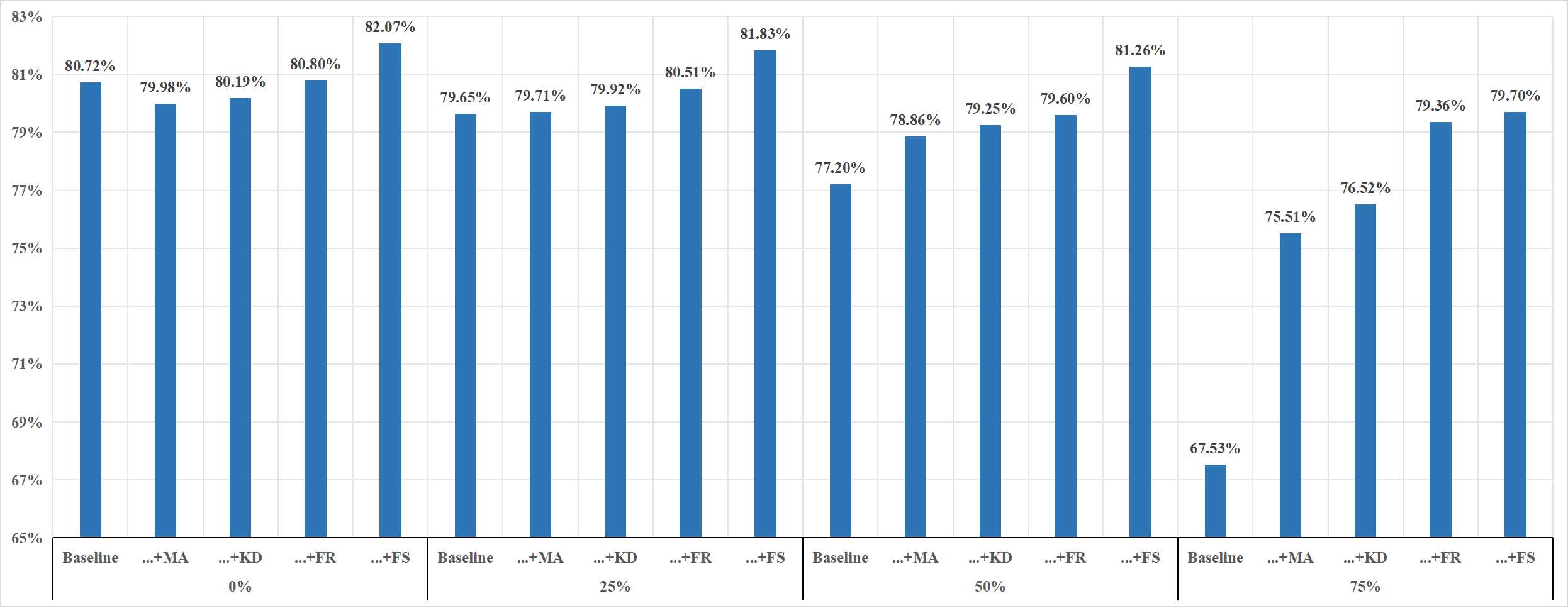}
		\caption{Performance of different module combinations on cropland F1-score under varying temporal missing conditions on Hunan SEN datasets. (MA, KD, FR, and FS represent the Mask, Distill, Feature Reconstruction, and Feature Standardization modules, respectively. "$\cdots$" indicates previously added modules.)}
		\label{F8}
	\end{center}
\end{figure}

As shown in Fig. \ref{F8}, with the increase of temporal missingness, the benefits of the MA module also grow. Although it shows a negative effect on complete SITS (0\% ), it plays a crucial role under more severe missing conditions, especially under high-level (75\%) missing conditions. Secondly, within the 0-50\% temporal missing range, the addition of KD and FR modules yields only limited gains compared to using MA alone. The incorporation of the FS module amplifies the effect of the FR module. However, under the high-level missing condition of 75\%, the trend is different. The addition of the FR module exhibits a significant improvement compared to the combination of MA and KD, while the FS module only provides limited gains. This may be because changes in feature magnitude also play an important role under severe temporal missingness, and excessive focus on the feature’s distribution might diminish some of the potential benefits. The interaction between FR and FS under different types and levels of temporal missing conditions still requires further investigation.

\begin{figure}[h!]
	\begin{center}
        \includegraphics[width=1\linewidth]{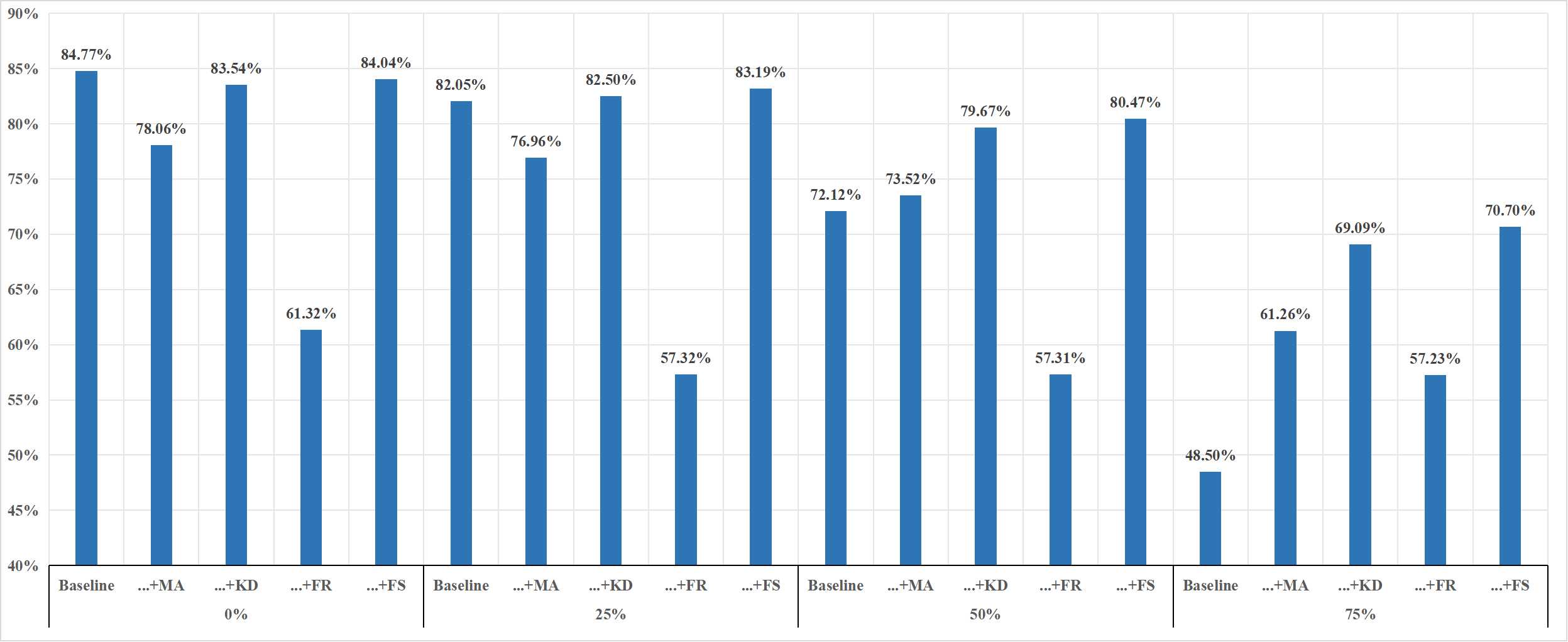}
		\caption{Performance of different module combinations on mean F1-scores under varying temporal missing conditions on Fr\&Cat S4A datasets. (MA, KD, FR, and FS represent the Mask, Distill, Feature Reconstruction, and Feature Standardization modules, respectively. "$\cdots$" indicates previously added modules.)}
		\label{F9}
	\end{center}
\end{figure}

As shown in Fig. \ref{F9}, the experimental trend in the crop classification task differs significantly from those observed in the cropland classification task. Firstly, without the FS module, the FR module causes severe overfitting in the model, resulting in a loss of classification ability for certain classes. This performance is even worse than using only the MA module under 0\%-50\% temporal missing conditions. This indicates that, without feature standardization, the reconstructed features may contain significant errors. Secondly, although the addition of the FR and FS modules improves the model’s performance, their gains are smaller compared to those brought by the KD module. This is because the soft supervised signal from the KD module provides the intrinsic relationships among different classes, which significantly reduces the risk of overfitting. We believe that in the crop classification task, alleviating model overfitting may play a more crucial role, while feature reconstruction also serves as a complementary enhancement.

\bibliographystyle{elsarticle-num} 
\bibliography{bib_MD_2_22.bib}

\end{document}